\documentclass[letterpaper,twocolumn,10pt]{article}
\usepackage{usenix}

\usepackage{tikz}
\usepackage{amsmath}
\usepackage{tabularx}
\usepackage{booktabs}
\usepackage{amsfonts}       
\usepackage{nicefrac}       
\usepackage{microtype}      
\usepackage{xcolor}         
\usepackage{tabularx}
\usepackage{booktabs}
\usepackage{makecell}
\usepackage{array}
\usepackage{comment}
\usepackage{graphicx}
\usepackage{subcaption}
\usepackage{enumitem}
\usepackage{adjustbox}
\usepackage{multirow}
\usepackage{url}
\usepackage{bm}

\begin{document}

\date{}

\title{LoMime: Query-Efficient Membership Inference using Model Extraction in Label-Only Settings}


\author{
{\rm Abdullah Caglar Oksuz}\\
Case Western Reserve University\\
abdullahcaglar.oksuz@case.edu
\and
{\rm Anisa Halimi}\\
IBM Research\\
anisa.halimi@ibm.com
\and
{\rm Erman Ayday}\\
Case Western Reserve University\\
erman.ayday@case.edu
}

\maketitle

\begin{abstract}
Membership inference attacks (MIAs) threaten the privacy of machine learning
models by revealing whether a data point was used during training. Existing
MIAs often assume access to public datasets, shadow models, confidence scores
or the training distribution, which makes them vulnerable to defenses like
confidence masking. Label-only MIAs avoid these assumptions but require
thousands of queries per sample. We propose a cost-effective label-only MIA
framework based on transferability and model extraction. Querying the target
$M$ with active sampling, perturbation-based selection and synthetic data, we
extract a surrogate $S$ on which membership inference is performed offline.
This shifts query overhead to a one-time extraction phase. It also removes the
restriction that defines the label-only setting: the attacker controls $S$ and
can read its posteriors and training trajectory, so attacks that cannot be run
against $M$ can be run against $S$. On Location, Purchase and Texas, the
strongest attack on $S$ improves AUC over the direct attack on $M$ by $0.9$,
$5.6$ and $5.0$ percentage points, and improves the true positive rate at $1\%$
false positive rate by $2.8\times$ to $6.3\times$. We characterize how leakage
transfer depends on surrogate fidelity, evaluate standard defenses, and report
preliminary results on image datasets.
\end{abstract}

\section{Introduction}
\label{sec:intro}

Machine learning (ML) models are increasingly deployed in sensitive, high-impact domains such as healthcare~\cite{healthcare} and finance~\cite{finance}. These deployments routinely rely on training corpora that encode sensitive attributes (e.g., medical history, financial transactions, or behavioral traces). When models are exposed via public APIs or interactive services, attackers can probe them remotely, and privacy risks emerge even when the service reveals only the final predicted label.

A central privacy threat in this setting is the \emph{membership inference attack} (MIA), in which an attacker aims to determine whether a particular sample was included in a model's training set~\cite{shokri2017membership}. Successful membership inference can enable consequential disclosures without requiring explicit reconstruction of training data (e.g., inferring participation in a disease-specific clinical trial). Conceptually, MIAs exploit the fact that many models behave differently on training points versus unseen points due to overfitting or memorization~\cite{shokri2017membership}.

\textbf{Label-only MIAs and the scalability bottleneck:} Recent work has shown that membership leakage can persist even under label-only black-box access, where the attacker observes only the hard label output~\cite{choquette, li_and_zhang}. A prominent class of label-only MIAs estimates how stable a model's predicted label is under perturbations of the input: members tend to exhibit different robustness patterns than non-members in overfit regimes~\cite{choquette}. While such attacks are attractive because they avoid unrealistic assumptions (e.g., confidence vectors or internal gradients), their main practical limitation is query inefficiency. Determining the membership of a single candidate point can require thousands of queries to approximate decision-boundary proximity or robustness statistics, and additional queries are often needed to calibrate an appropriate decision-boundary distance threshold $\tau$ for the target model~\cite{choquette, li_and_zhang}. Since each candidate is paid for separately against the live model, these per-sample costs accumulate linearly with the number of samples to be tested, and they hinder the use of label-only MIAs for large-scale auditing when APIs enforce rate limits, monetization, or anomaly detection.

\textbf{Our approach: amortizing label-only membership inference via model extraction:} We propose a cost-effective label-only membership inference framework that shifts expensive, repeated queries into a one-time extraction phase. Instead of querying the target model $M$ independently for each membership decision, we first extract a surrogate model $S$ that approximates $M$ under hard-label feedback, and then perform all membership inference offline against $S$ using the same label-only robustness principles. Extraction, however, does more than relocate the query cost. The attacker trains $S$ and holds its parameters, so its posteriors, gradients and training trajectory are available even though $M$ never exposed anything beyond a hard label. A label-only threat model therefore becomes a white-box one on the surrogate, and attacks that cannot be run against $M$ can be run against $S$. We exploit this by attacking $S$ with posterior-based metric attacks and with SeqMIA~\cite{seqmia}, a sequential-metric attack over the surrogate's distillation trajectory, alongside the label-only boundary-distance attack. This reframes the role of model extraction in privacy analysis: prior work typically treats membership inference as a probe of how informative an extracted model is~\cite{marich, nasr}, whereas we use extraction as the mechanism that amortizes the query cost of the attack itself. The strategy leverages the empirical transferability of decision-boundary behavior across functionally similar models~\cite{papernot2016transferability, yanpeiliu, naseer, demontis}, and it makes the marginal cost of additional membership queries essentially zero after extraction. Thus, once $S$ is obtained, membership can be audited at a scale that direct label-only MIAs cannot reach.

\textbf{Extraction under strict, pool-free constraints:} Our extraction stage builds on label-only active learning, in which informative queries are selected to train a surrogate that agrees with the target. The closest precedent, MARICH~\cite{marich}, selects these queries from a public query pool, an assumption central to its design but often unavailable in tabular settings, or whose provenance is itself sensitive. We therefore remove the public-pool dependency for tabular data: starting from only a small seed set, we synthesize candidate queries through replication and structured probabilistic perturbation, adopting the perturbation strategy (though not the explanation-guided feature selection) of AUTOLYCUS~\cite{autolycus}. Generating queries that are informative enough to drive active sampling without a curated pool is the central extraction-side challenge we address. For high-dimensional image data, where synthesizing meaningful queries in pixel space is substantially harder, we retain a public out-of-distribution pool and state this boundary explicitly (Section~\ref{sec:dnn_extension}). In all cases the attacker uses only hard labels and assumes no knowledge of the target's architecture, hyperparameters, training data, or training distribution. The seed set is small and serves only to bootstrap query synthesis rather than to provide broad distributional coverage; we analyze its sensitivity to size and distribution in Section~\ref{sec:eval}.

\textbf{Contributions:} Our main contributions are as follows:
\begin{itemize}[leftmargin=*]
    \item \textbf{Extraction as a query-cost amortizer for label-only MIA.} We recast model extraction as a mechanism that converts the per-sample online query cost of label-only membership inference into a one-time, offline cost, enabling membership auditing at a scale that direct label-only attacks cannot achieve under realistic query constraints.

    \item \textbf{Pool-free label-only extraction.} We remove the public query-pool assumption of prior label-only extraction~\cite{marich} in the tabular setting and replace it with synthetic query generation from a small seed via replication and structured perturbation. We retain a public out-of-distribution pool only for the high-dimensional image setting.

    \item \textbf{From label-only to white-box via extraction.} The surrogate is not only a cheaper copy of the target, it is also a less restricted attack surface. The attacker controls $S$, so posterior- and trajectory-based attacks that the hard-label interface of $M$ rules out become applicable. Attacking $S$ this way exceeds the label-only attack run directly on $M$ in AUC on all three tabular benchmarks, and by a larger margin at low false positive rates.

    \item \textbf{When amortization pays off, and how leakage transfers.} We provide an explicit break-even characterization of the query regime in which extraction-based auditing is cheaper than direct label-only MIA. Moreover, we analyze how membership-leakage transfer depends on surrogate fidelity, including regimes where high fidelity does not by itself imply strong leakage transfer.

    \item \textbf{Evaluation under defenses.} We evaluate the framework on standard benchmark datasets and study common countermeasures proposed for label-only MIAs, including dropout~\cite{dropout}, L2 regularization~\cite{andrewng, l2}, and DP-SGD~\cite{dpsgd}. Unless strong differential privacy is applied, the surrogate retains enough leakage signal for effective offline membership inference.
\end{itemize}

\textbf{Paper organization:} Section~\ref{sec:relatedwork} reviews prior work on membership inference and model extraction. Section~\ref{sec:systhreatmodel} specifies the system and threat model. Section~\ref{sec:methodology} presents our methodology. Section~\ref{sec:eval} details the experimental setup and results. Section~\ref{sec:countermeasures} evaluates defenses. Section~\ref{sec:dnn_extension} discusses the potential of extending the proposed framework to deep neural networks, and Section~\ref{sec:conclusion} concludes and outlines future directions.

\section{Related Work}
\label{sec:relatedwork}

\textbf{Membership Inference Attacks:} Early MIAs typically operate in confidence-based or white-box settings, and often assume access to auxiliary datasets and/or surrogate models to learn model-specific membership signals~\cite{shokri2017membership, salem2018ml, pyrgelis, truex, hayes, hilprecht, song, sablayrolles, long, jiachengli, hui, nasr, memguard, yang}. These assumptions can be unrealistic in production APIs, where confidence vectors may be masked, quantized, or withheld entirely. Accordingly, a parallel line of work has studied defenses such as confidence masking~\cite{shokri2017membership, salem2018ml, truex, memguard, yang}, adversarial regularization~\cite{nasr}, and techniques that improve generalization~\cite{shokri2017membership, salem2018ml, truex, dpsgd, dropout}.

More recent research has pursued diverse directions, including reducing attack cost~\cite{zarifzadeh}, developing MIAs for large language models~\cite{duan, mireshghallah, mattern}, leveraging explainability for membership signals~\cite{mia_explainability}, using MIAs for privacy auditing~\cite{jiayuan_ye}, and optimizing performance in low false-positive regimes~\cite{carlini_lira}. In the label-only setting, Yeom et al.~\cite{yeom} proposed a simple baseline that flags correctly classified points as members, and subsequent work showed that perturbation-based robustness probing can recover stronger membership signals under hard-label access~\cite{choquette, li_and_zhang}. Our work focuses on this strict label-only regime and addresses its main bottleneck: query inefficiency when attacks are run directly against the deployed model.

\textbf{Recent label-only attacks and their assumptions:} A recent line of work attacks the same bottleneck by reducing the per-sample query cost. YOQO~\cite{yoqo} crafts a single query sample whose hard label reveals membership, cutting the budget to one query per candidate. OSLO~\cite{oslo} constructs a transfer-based adversarial example calibrated so that the perturbation flips a non-member but not a member, again using a single query. DHAttack~\cite{enhanced-lomia} replaces the shortest boundary distance with the distance measured toward a fixed out-of-distribution point and calibrates it against a Gaussian model of the sample's non-member state, reaching strong TPR at low FPR with 5 to 30 queries per candidate. These attacks are effective, but they relocate the cost rather than remove it, and they do so by strengthening the adversary's assumptions in ways our threat model does not permit. Each requires shadow or surrogate models trained on data drawn from the target's training distribution, and DHAttack further assumes the target's architecture and hyperparameters are known. We detail these assumptions, and compare against the results they report on the tabular benchmarks we use, in Section~\ref{sec:comparison}.

\textbf{Model Extraction Attacks:} Model extraction attacks aim to replicate a target model's functionality by issuing queries and training a surrogate that matches the target's behavior~\cite{tramer, orekondy2019knockoff, prada, papernot, jagielski, truong, krishna, marich}. Depending on what the API reveals (probabilities, logits, or only labels), extraction strategies range from straightforward distillation to active learning and data-free synthesis. Hard-label extraction is especially challenging due to limited feedback, but can remain effective with adaptive query selection and informative sampling~\cite{marich}. MARICH~\cite{marich} is particularly relevant to our setting: it is label-only and emphasizes query efficiency through active sampling guided by a mutual-information objective. Specifically, MARICH composes existing active-learning samplers (entropy, entropy-gradient, and loss sampling) into a sequential selection pipeline over a public query pool. In MARICH this pool (possibly out-of-distribution) is assumed to be available, whereas for tabular data we target deployments where such a pool is unavailable or its provenance is itself sensitive, and we therefore replace it with synthetic query generation and structured perturbation; for high-dimensional image data we retain an out-of-distribution pool (Section~\ref{sec:dnn_extension}).

\section{System and Threat Model}
\label{sec:systhreatmodel}

\begin{figure*}[ht]
\centering
\includegraphics[width=\textwidth]{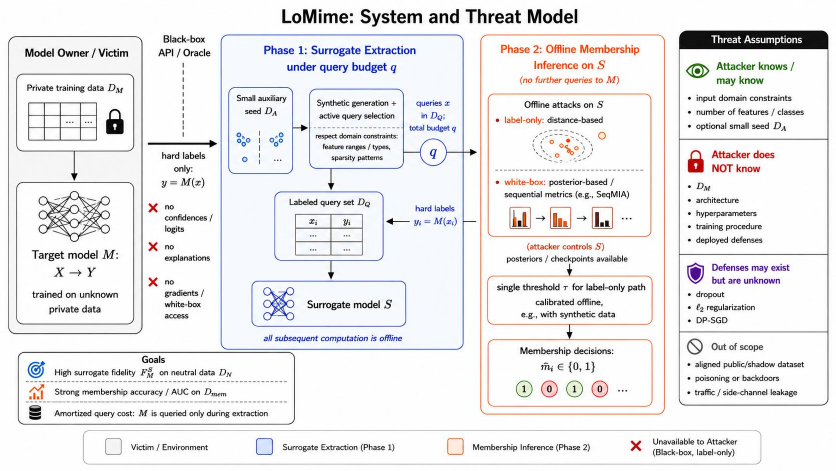}
\caption{Overview of the LoMime system and threat model. Phase~1 extracts a surrogate $S$ of the black-box target $M$ under query budget $q$; the oracle returns only hard labels. Phase~2 performs membership inference offline on $S$, by the label-only path that would also apply to $M$, or by the white-box path.}
\label{fig:threatmodel}
\end{figure*}

Figure~\ref{fig:threatmodel} summarizes our system and threat model. We consider a deployed black-box classifier $M:\mathcal{X}\to\mathcal{Y}$, trained on an unknown dataset $D_M$, as the target model. The attacker interacts with $M$ only through an oracle that returns hard labels $y_i=M(x_i)$ for inputs $x_i\in\mathcal{X}$, and trains a surrogate $S$ from this feedback alone. A finite query budget $q$ limits the total interaction with $M$ during extraction; all subsequent computations on $S$ are offline and unconstrained by $q$. The attacker has no access to $D_M$, to a curated public or shadow dataset aligned with $M$'s training distribution, or to any prior knowledge of $M$'s architecture, hyperparameters, or defenses. The only data the attacker holds is a small seed set $D_A$, which bootstraps query synthesis rather than providing broad distributional coverage. Crucially, $D_A$ is a convenience rather than a hard requirement: because each round of queries to $M$ progressively supplies the supervision that drives extraction, the influence of $D_A$ is concentrated in the early, low-budget rounds and shrinks as the budget grows. The attack therefore remains effective with substantially smaller seeds, and the size and provenance of $D_A$ chiefly govern how quickly the surrogate converges rather than the fidelity and membership performance it ultimately attains. We do not assume that $D_A$ matches the target's training distribution; in our tabular experiments we instantiate it as a disjoint split of the same dataset at roughly $10\%$ of $|D_M|$ for comparability with prior work, and we quantify its size and distribution sensitivity, including the near data-free setting, in Section~\ref{sec:eval}. Generated inputs must satisfy domain constraints (e.g., feature ranges, types, and sparsity patterns), and explanations or other side channels are never provided. Defenses such as dropout, $\ell_2$ regularization, or DP-SGD may be present, but they are unknown to the attacker and treated as part of the environment. This strict, pool-free model describes the tabular setting; the deep-learning extension in Section~\ref{sec:dnn_extension} relaxes the no-pool assumption by permitting an out-of-distribution public pool, since synthesizing informative queries directly in pixel space is substantially harder. Appendix~\ref{app:notation} summarizes the symbols and notation used throughout.

In the extraction phase, the attacker's objective is to learn a surrogate $S$ that closely agrees with $M$, measured by label-agreement fidelity $F_{M}^{S}=\Pr_{x\sim\mathcal{D}_N}[S(x)=M(x)]$ on a neutral test set $D_N$ under budget $q$. In the membership inference phase, given an evaluation set $D_{\mathrm{mem}}=\{x_i\}$, the objective is to infer membership bits $\hat m_i\in\{0,1\}$ that indicate whether each $x_i$ belongs to $D_M$, with performance reported via accuracy and AUC when ground truth is available. A key property of this model is the amortization of query cost: all queries to $M$ occur once, during extraction, after which membership inference is performed offline on $S$ without further access to $M$. Under the label-only attack, each membership decision is derived from a scalar distance-based score computed on $S$ and a single threshold $\tau$, calibrated in an unsupervised manner from synthetic samples and without querying $M$. The white-box attacks of Section~\ref{sec:whitebox} score $S$ directly from its posteriors and distillation trajectory, again without any further access to $M$. Accordingly, our threat model excludes confidence or logit outputs, explanations, white-box or gradient interfaces, curated public or shadow datasets aligned with $D_M$, poisoning or backdoor manipulations, and traffic or side-channel leakage. These exclusions apply to the target $M$, which is the only model the attacker does not control. They do not apply to the surrogate. $S$ is trained by the attacker, so its posteriors, gradients and intermediate checkpoints are available, and withholding confidences is a defense the deployer can apply to $M$ but not to a model the attacker holds. Section~\ref{sec:whitebox} uses this asymmetry. The aim is to achieve high $F_{M}^{S}$ and competitive test performance under budget $q$, by attaining strong membership accuracy and AUC on $D_{\mathrm{mem}}$. Finally, for comparability with prior membership-inference work, our target models adopt the dataset and training configurations established in that literature, which we detail in Section~\ref{sec:eval}.

\section{Methodology}
\label{sec:methodology}

Our methodology consists of two steps: The first stage of our framework is model extraction, and the second stage is offline membership inference. Since the target model $M$ exposes only labels, we require a surrogate-construction procedure that is both query-efficient and faithful, in the sense of matching $M$'s predictions. 

\subsection{Model Extraction Attack}

In active learning, extraction is framed as selecting, in each iteration, the queries that most improve the surrogate's agreement with the target. In iteration $t$, let $D_Q^t$ denote the candidate query set obtained by augmenting and perturbing $D_A$ (described below), and let $Q \subseteq D_Q^t$ be a batch of queries drawn from it. Writing $\Pr(M(Q))$ and $\Pr(S(Q))$ for the label distributions that the target and the surrogate induce over $Q$, each iteration seeks a surrogate whose predictions are maximally informative about, and in agreement with, the target's. MARICH formalizes this as maximizing the mutual information between the two prediction distributions:
\begin{equation}
\label{eq:objective}
\max_{S}\; I\big(\Pr(M(Q)),\,\Pr(S(Q))\big), \qquad Q \sim D_Q^t .
\end{equation}
Here the maximization is over the surrogate $S$ (its parameters). Intuitively, the objective reduces the mismatch between the prediction distributions while keeping the surrogate's predictions informative, so that each queried label is as useful as possible~\cite{marich}.

The extraction proceeds in five iterative steps: (1) augmenting and perturbing $D_A$ to form candidates (AUTOLYCUS-style), then (2) entropy, (3) entropy-gradient, and (4) loss sampling to select the most informative queries from the candidate queries $D_Q$ (MARICH-style), and finally (5) updating the surrogate on the accumulated labeled set. The update warm-starts from the previous iteration's surrogate rather than retraining from scratch.

Extraction begins by training an initial surrogate $S^0$ on $D_A$ labeled by $M$. The seed $D_A$ is fixed throughout; at each iteration $t$ we regenerate a fresh candidate set $D_Q^t$ from it by replicating $D_A$ a factor of $r$ times, so that $|D_Q^t| = r\,|D_A|$ has the same nominal size in every iteration. Each of the $r\,|D_A|$ copies then receives a unique perturbation mask that alters each feature with probability $\rho$: for binary features the mask applies Bernoulli flips, while for continuous features it adds noise scaled by the per-feature standard deviation $\sigma_j$ (or a user-set value), always keeping the feature within its valid range $R_j$. Because the masks are drawn afresh each iteration, the candidates differ across rounds; we additionally discard any sample that duplicates an existing surrogate training point in $D_S^{t-1}$ before passing $D_Q^t$ to entropy sampling.

In the entropy sampling stage, a subset $Q_{\text{entropy}}^t$ of size $|Q_{\text{entropy}}^t| = B$ is selected from $D_Q^t$ by maximizing the entropy of the surrogate's predictions, as in Equation~\ref{eq:entropy}, where $H(\cdot)$ is the entropy and $B$ is the per-iteration sampling budget (its value is given in Section~\ref{sec:eval}). The subset $Q_{\text{entropy}}^t \subseteq D_Q^t$ marks the inputs about which $S^{t-1}$ is least certain.
\begin{equation}
\label{eq:entropy}
Q_{\text{entropy}}^t = \arg\max_{Q \subseteq D_Q^t,\, |Q| = B} H\big(S^{t-1}(Q)\big),
\end{equation}
In the entropy-gradient stage, $Q_{\text{entropy}}^t$ is refined by clustering the entropy gradients $\nabla_Q H(S^{t-1}(Q))$ of the selected queries with $k$-means, using $k = n_c$ clusters so that each label is represented. From these clusters, the most diverse subset $Q_{\text{grad}}^t$ of size $\gamma_1 B$ is chosen to maximize input-space coverage (Equation~\ref{eq:gradient}), where $C$ denotes the cluster centers.
\begin{equation}
\label{eq:gradient}
Q_{\text{grad}}^t = \arg\min_{Q \subseteq Q_{\text{entropy}}^t,\, |Q| = \gamma_1 B} \sum_{x_i \in Q} \sum_{c \in C} \|\nabla_Q H(S^{t-1}(x_i)) - c\|^2 .
\end{equation}
The stage targets balanced label representation, but it is not guaranteed: when some clusters are underpopulated, for instance for rare labels or sparse feature regions, the shortfall is filled from the remaining clusters so that $\gamma_1 B$ points are still returned, at the cost of uneven coverage for large $n_c$. This partly accounts for the slower convergence we observe on high-cardinality datasets (Section~\ref{sec:eval}).

In the loss sampling stage, $Q_{\text{grad}}^t$ is further refined to $\gamma_1 \gamma_2 B$ queries $Q_{\text{loss}}^t$, namely the candidates closest to the surrogate training points on which the target and the surrogate disagree most. Concretely, for each $x \in D_S^{t-1}$ we compute the cross-entropy loss of $S^{t-1}$ under the label $y$ provided by $M$, $L(x) = -\log p_{S^{t-1}}(y \mid x)$, and collect the highest-loss training points as $D_{S,\text{top}}^{t-1}$. We then select the $\gamma_1 \gamma_2 B$ candidates in $Q_{\text{grad}}^t$ nearest to $D_{S,\text{top}}^{t-1}$ (Equation~\ref{eq:loss}).
\begin{equation}
\label{eq:loss}
Q_{\text{loss}}^t = \arg\min_{Q \subseteq Q_{\text{grad}}^t,\, |Q| = \gamma_1 \gamma_2 B} \sum_{x_i \in Q} \sum_{s \in D_{S,\text{top}}^{t-1}} \|x_i - s\|^2 .
\end{equation}
Only $Q_{\text{loss}}^t$ is sent to $M$ for labels $Y_{\text{loss}}^t = M(Q_{\text{loss}}^t)$, so each iteration costs $\gamma_1 \gamma_2 B$ queries and the total budget is $q = \sum_t \gamma_1 \gamma_2 B$. The labeled samples extend the training set as $D_S^t = D_S^{t-1} \cup Q_{\text{loss}}^t$ and $Y_S^t = Y_S^{t-1} \cup Y_{\text{loss}}^t$. The surrogate is then updated on the extended set, $S^t = \text{train}(D_S^t, Y_S^t)$, warm-started from $S^{t-1}$ and optimized for $E$ epochs with SGD or Adam~\cite{adam}. Steps (1)--(5) repeat until the budget $q$ is exhausted or a target fidelity $F_M^S$ is reached, yielding the final surrogate $S$. This procedure ensures high fidelity, informativeness, and diversity while remaining query-efficient.

\subsection{Label-Only Membership Inference Attack}

Extracting a functionally equivalent surrogate $S$ lets the attacker perform membership inference offline on a dataset $D_{\text{mem}}$ (whose membership we aim to determine), circumventing the high per-sample query costs of direct label-only attacks. It also removes unrealistic assumptions such as shadow datasets and large numbers of online adversarial queries, which both traditional and label-only MIAs typically require. Label-only MIAs match traditional MIAs when query budgets are unconstrained~\cite{choquette}. To simulate a setting where $S$ may originate from an external resource or an open repository with no access to $D_S$, and where no data from $M$'s training distribution is available, we conduct an unsupervised label-only MIA on $S$ in the second stage.

For each candidate $x_i \in D_{\text{mem}}$, the attack uses only the hard-label prediction $\hat{y}_i = S(x_i)$ together with the proximity of $x_i$ to the decision boundary of $S$. We define the boundary distance as the smallest perturbation that changes the surrogate's prediction:
\begin{equation}
\label{eq:distance}
d_{\text{boundary}}(x_i) = \min_{\delta_i \in \mathbb{R}^d} \| \delta_i \|_2 \quad \text{subject to} \quad S(x_i + \delta_i) \neq S(x_i) .
\end{equation}
Membership is then decided by a single threshold $\tau$,
\begin{equation}
\label{eq:membership}
\hat{m}_i =
\begin{cases}
1 & \text{if } d_{\text{boundary}}(x_i) \geq \tau, \\
0 & \text{if } d_{\text{boundary}}(x_i) < \tau,
\end{cases}
\end{equation}
where $\hat{m}_i$ is the predicted membership label. The intuition is that training members tend to lie farther from the decision boundary than non-members because of overfitting, so a larger boundary distance is evidence of membership.

We calibrate $\tau$ on $S$ without querying $M$. Following the label-only boundary attack of Li and Zhang~\cite{li_and_zhang}, we draw a set of synthetic calibration samples $X_{\text{cal}}$ (uniform and out-of-distribution points drawn within the valid feature ranges and disjoint from $D_S$), compute the boundary distance $d_{\text{boundary}}(x_i)$ from Equation~\ref{eq:distance} for each, and set $\tau$ to the $p$-th percentile of the resulting distances:
\begin{equation}
\label{eq:tau}
\tau = \mathrm{Percentile}_{p}\big(\{\, d_{\text{boundary}}(x_i) : x_i \in X_{\text{cal}} \,\}\big).
\end{equation}
Li and Zhang report that effective thresholds fall in the 30th--80th percentile range, and we use the median ($p = 50$) as a robust default. Unlike a maximum or other extreme statistic, a percentile threshold is insensitive to outliers in the calibration set. The scalar threshold can be generalized to class-specific thresholds $T = \{\tau_1, \tau_2, \ldots, \tau_{n_c}\}$ when auxiliary information is available (e.g., decision regions, population distributions, or partial training data). However, because decision regions in complex models are highly irregular and nonlinear and the attacker has little or no such information, we adopt a single global threshold.

\subsection{White-Box Inference on the Surrogate}
\label{sec:whitebox}

The attack described above treats $S$ exactly as it would treat $M$: it observes only hard labels and infers membership from boundary distance. We use it as the baseline because it isolates the effect of amortization from any change in the attack itself. It is, however, more restrictive than the setting requires. The attacker trained $S$ and holds its parameters, so its output posteriors, its gradients and every intermediate checkpoint of its training are available. We therefore evaluate two further families of attacks on $S$, neither of which can be applied to $M$ under our threat model.

\textbf{Posterior-based metric attacks.} For each candidate $x_i \in D_{\text{mem}}$ we compute standard membership metrics from the surrogate's output posterior $S(x_i)$: the cross-entropy loss under the ground-truth label, the maximum posterior probability, the standard deviation of the posterior vector, and the prediction entropy. Each metric serves directly as a membership score, so these attacks need no additional models and no threshold calibration for the threshold-free metrics we report.

\textbf{Sequential-metric attack.} We also apply SeqMIA~\cite{seqmia} to $S$. SeqMIA distills the model under attack, snapshots the distilled student at every epoch, evaluates the metrics above on each snapshot, and assembles the resulting $k \times (n+1)$ multi-metric sequence. An attention-based recurrent attack model then classifies membership from the sequence, exploiting the temporal pattern of the metrics rather than their final values alone. Applying SeqMIA to $S$ requires one adjustment. Since $S$ was only trained on synthetic queries labeled by $M$, its metric sequences carry $D_M$-membership only through the decision geometry it inherited from $M$. Training the attack model on $S$'s own training set would predict membership in $D_S$, which is not the quantity we want. We instead train it on a \emph{shadow extraction}: we train a local target $M'$ on data disjoint from $D_M$, extract $S'$ from $M'$ using the same procedure and budget that produced $S$ from $M$, and label the resulting sequences using $M'$'s known training membership. Every query in this pipeline is issued to $M'$, a model the attacker owns, so the shadow extraction costs computation but no additional queries to $M$. The distillation set is synthesized from the seed exactly as extraction queries are, so this stage introduces no data assumption beyond those already stated in Section~\ref{sec:systhreatmodel}.

\subsection{Performance Evaluation}

We assess the attack along two axes: extraction and membership inference. For extraction, we report (i) fidelity $F_M^S$ (label agreement between $S$ and $M$), (ii) test accuracy $\text{Acc}(\cdot,D_N)$ of $S$ and $M$, and (iii) the query budget $q$ consumed during extraction. For membership inference, we report (i) accuracy on $D_{\text{mem}}$ at a calibrated threshold, (ii) threshold-free AUC from continuous membership scores, and (iii) the true positive rate at fixed low false positive rates, TPR@$0.1\%$FPR and TPR@$1\%$FPR. The last of these follows Carlini et al.~\cite{carlini_lira}, who argue that average-case metrics understate the practical threat of an MIA and that an attack is only useful if it identifies some members while making few false accusations. We report both because they need not agree, and Section~\ref{sec:eval} contains a case where they rank different attacks.

High $F_M^S$ together with competitive $\text{Acc}(\cdot,D_N)$ at a modest $q$ indicates a successful, cost-effective extraction. Comparing membership metrics between $S$ and $M$ measures leakage transfer: close accuracy and AUC imply that $S$ preserves the target's membership signal, whereas a gap suggests incomplete transfer (e.g., underfitting during extraction). We analyze this fidelity-to-leakage relationship, including cases where high fidelity does not by itself yield strong transfer, in Section~\ref{sec:eval}. Since membership inference on $S$ is offline, the approach is advantageous at scale: once $S$ is extracted, a large $D_{\text{mem}}$ can be evaluated without any further queries to $M$.

\section{Evaluation}
\label{sec:eval}

In this section, we describe the datasets, experimental setup, sources of variability, computational settings, evaluation metrics, and the obtained results.

\subsection{Datasets}

We evaluate our framework using three benchmark datasets widely adopted in privacy and security research~\cite{shokri2017membership}: \textbf{Location}, \textbf{Purchase}, and \textbf{Texas Hospital}. The statistics of these datasets are summarized in Table~\ref{tab:dataset_config}. We train the target models following standard configurations in the membership inference literature~\cite{shokri2017membership, choquette} for direct comparability with prior label-only attacks. In benchmarks, these models are generally overfitted with near-perfect-training-accuracy.

\begin{table}[ht]
\centering
\caption{Dataset statistics}
\begin{adjustbox}{width=0.9\columnwidth}
\begin{tabular}{lcccccccc}
\toprule
\textbf{Dataset} 
& \makecell{\textbf{\# of Total} \\ \textbf{Samples} ($\mathbf{|D|}$)} 
& \makecell{\textbf{\# of Features} \\ ($\mathbf{n_j}$)} 
& \makecell{\textbf{\# of Classes} \\ ($\mathbf{n_c}$)} \\
\midrule
Location       & 5,010   & 446   & 30  \\
Purchase       & 197,324 & 600   & 100 \\
Texas Hospital & 67,330  & 6,170 & 100 \\
\bottomrule
\end{tabular}
\end{adjustbox}
\label{tab:dataset_config}
\end{table}

\subsection{Experimental Setup}
In our experiments, both target $M$ and surrogate models $S$ are implemented and trained using PyTorch. Following the setup of~\cite{shokri2017membership}, $M$ are configured as feedforward neural networks with a single hidden layer of 128 nodes and the tanh activation function. Models are trained for up to 200 epochs using the AdamW optimizer~\cite{adamw}, with a learning rate of 0.001 and a weight decay coefficient of $\lambda = 10^{-7}$. Batch shuffling is applied at each epoch, with batch sizes of 100 for the Location dataset and 200 for the Purchase and Texas Hospital datasets. To simulate the attacker’s limited knowledge of dataset characteristics, augmentation and perturbation parameters used across all datasets during model extraction are set to $k = n_j$ (uniform binary flipping), $r = 4$ (the seed replication factor of Section~\ref{sec:methodology}), and $\gamma_1 = \gamma_2 = 0.5$. Apart from $\rho$, introduced parameters are architecture agnostic in our framework and primarily regulate per-iteration training-set growth rather than extraction quality. If $\rho$ is too large, perturbations can yield unrealistic queries; if too small, queries become uninformative; both cases degrade extraction fidelity. Hence, a set of synthetic samples or the initial $D_A$ can be utilized to configure $\rho$ as per-feature standard deviation across the seed, or the nonzero-feature fraction. Table~\ref{tab:exp_config} presents the experimental configuration, including the sizes of the training and auxiliary datasets, the perturbation factor $\rho$, and the resulting target model accuracy. Appendix~\ref{app:var_comp} reports the number of repetitions behind the standard deviations we quote, and the hardware used.

\begin{table}[ht]
\centering
\caption{Training data composition, adversarial parameters, and target model performance}
\begin{adjustbox}{width=\columnwidth}
\begin{tabular}{lcccccccc}
\toprule
\textbf{Dataset} 
& \makecell{\textbf{\# of Training} \\ \textbf{Samples} ($\mathbf{|D_M|}$)} 
& \makecell{\textbf{\# of Auxiliary} \\ \textbf{Samples} ($\mathbf{|D_A|}$)} 
& \makecell{$\bm{\rho}$} 
& \makecell{$\mathbf{M}$ \textbf{Testing} \\ \textbf{Accuracy (\%)}} \\
\midrule
Location       & 1,600  & 150   & 0.10  & 60.33 $\pm$ 0.84 \\
Purchase       & 10,000 & 1,000 & 0.08  & 64.89 $\pm$ 0.31 \\
Texas Hospital & 10,000 & 1,000 & 0.005 & 48.19 $\pm$ 0.24 \\
\bottomrule
\end{tabular}
\end{adjustbox}
\label{tab:exp_config}
\end{table}

We use IBM’s \textbf{Adversarial Robustness Toolbox (ART)}~\cite{art_toolbox} to implement standardized label-only membership inference attacks~\cite{choquette, li_and_zhang}. The label-only attacks on $M$ are then used as our benchmarks and our goal is to match their performance using significantly fewer queries. 

\subsection{Experiments}

\begin{figure*}[t]
    \centering
    \begin{subfigure}[t]{0.32\textwidth}
        \centering
        \includegraphics[width=\textwidth]{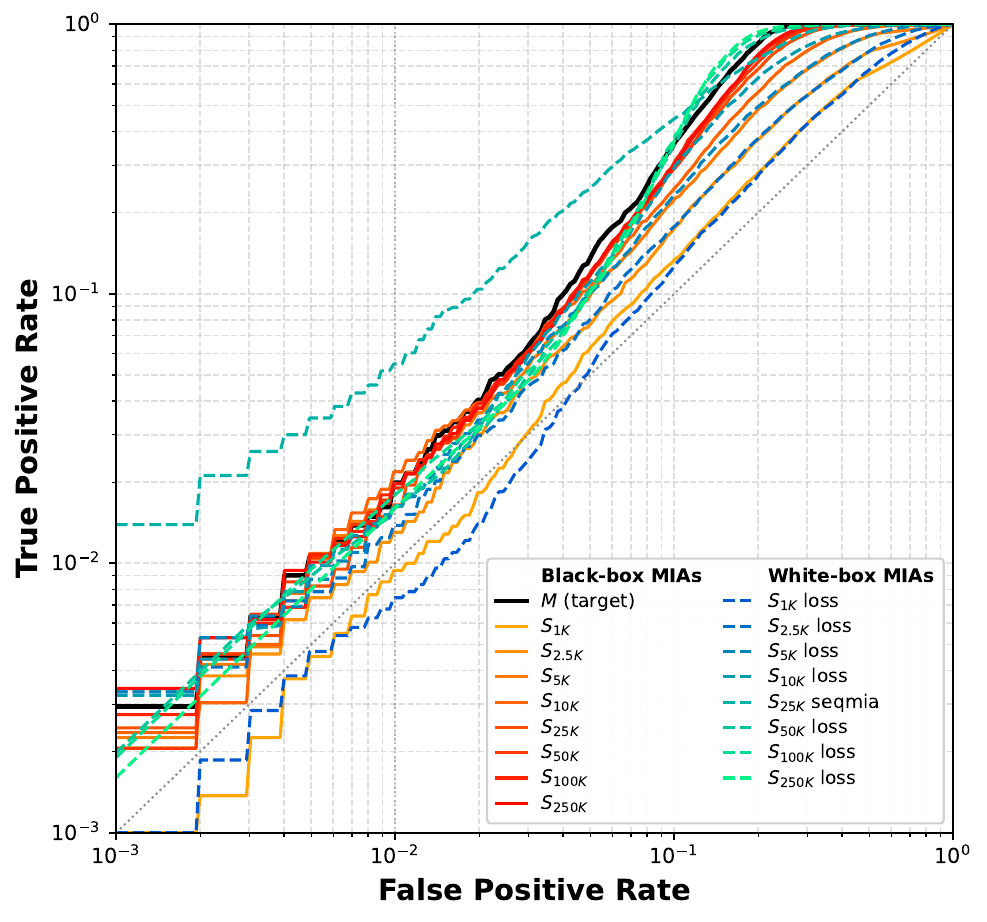}
        \caption{Location}
        \label{fig:roc_location}
    \end{subfigure}
    \hfill
    \begin{subfigure}[t]{0.32\textwidth}
        \centering
        \includegraphics[width=\textwidth]{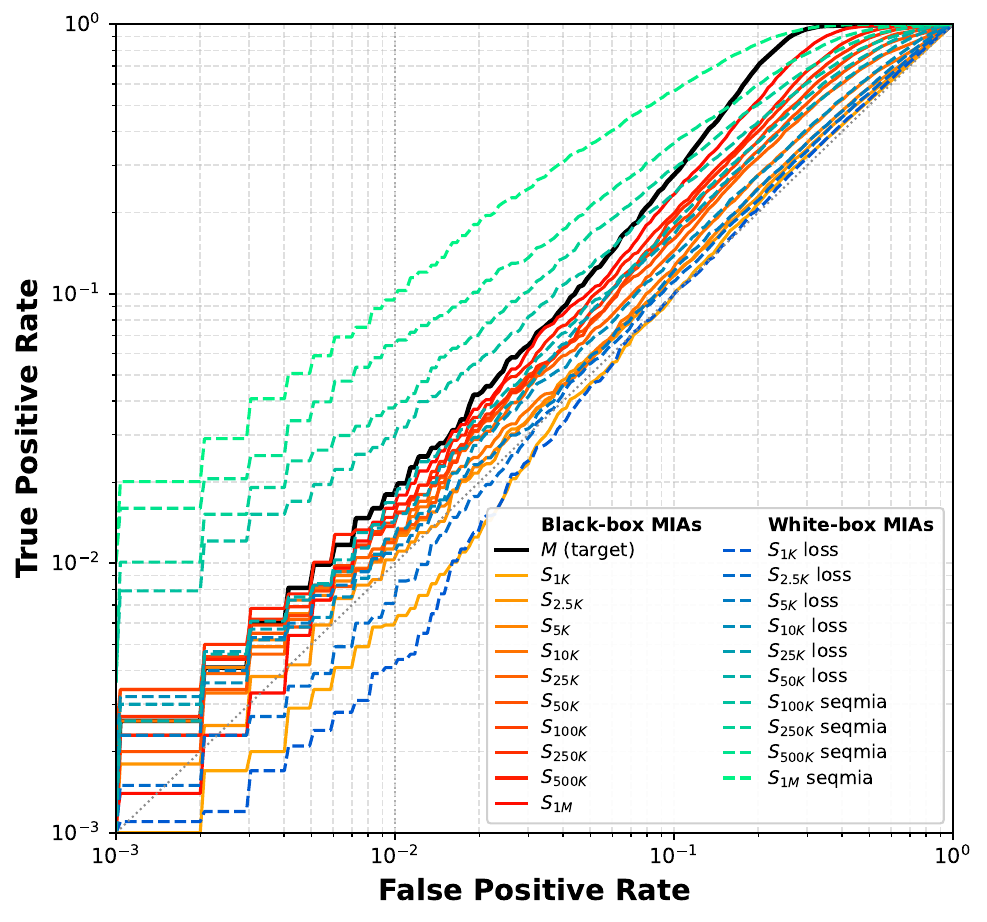}
        \caption{Purchase}
        \label{fig:roc_purchase}
    \end{subfigure}
    \hfill
    \begin{subfigure}[t]{0.32\textwidth}
        \centering
        \includegraphics[width=\textwidth]{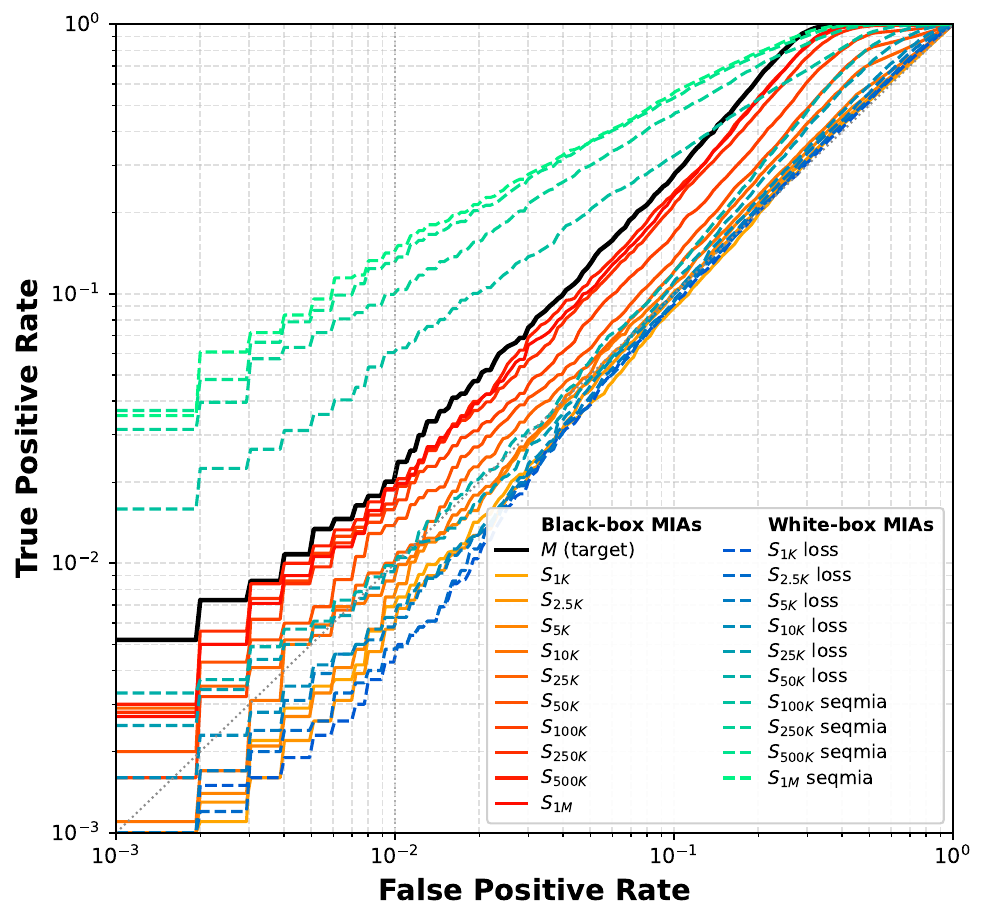}
        \caption{Texas}
        \label{fig:roc_texas}
    \end{subfigure}
    \caption{Log-scale ROC curves for membership inference on surrogate and target models across datasets. The logarithmic axes make the low false-positive region visible. Linear-axis versions are in Appendix~\ref{app:roc_linear}.}
    \label{fig:roc_combined}
\end{figure*}

We assess the effectiveness of our attacks using five key metrics, grouped into two categories: \textbf{model extraction metrics} and \textbf{membership inference metrics}. The first three (\textbf{fidelity of the surrogate model}, \textbf{test accuracy}, and \textbf{the query budget}) primarily evaluate the model extraction stage. Fidelity ($F_{\text{M}}^{\text{S}}$) quantifies the label agreement between $S$ and $M$ over a neutral dataset ($D_N$), which measures the percentage of predictions where $S$ and $M$ produce identical outputs, indicating how accurately $S$ replicates $M$'s decision boundaries. Test accuracy reflects the classification performance of both the surrogate model ($S$) and the target model ($M$) on $D_N$ and its ground-truth labels, providing a standardized evaluation of their generalization capabilities. The query budget determines the number of samples allowed to be classified by $M$. The remaining metrics, \textbf{attack accuracy}, the area under the ROC curve (\textbf{AUC}), and the \textbf{true positive rate at fixed low false positive rates} (TPR@$0.1\%$FPR and TPR@$1\%$FPR), are used to assess membership inference effectiveness, based on false positive rates (FPR) and true positive rates (TPR). While each metric primarily supports its corresponding attack component, they also offer insight into the relationship between extraction fidelity and membership leakage.

We further evaluate the model extraction process by measuring how well the surrogate model $S$ replicates the membership inference performance of the target model $M$ in a black-box, label-only setting. We evaluate the attack across surrogate models which have varying query budgets $q$. For the Location dataset we sweep $q$ from $1\text{K}$ to $250\text{K}$, and for the Purchase and Texas datasets from $1\text{K}$ to $1\text{M}$. Accordingly, a model denoted as $S_{\text{100K}}$ represents a surrogate trained by querying the target model 100{,}000 times. Our extraction and inference performances across a representative subset of these budgets are provided in Table~\ref{tab:query_budget_summary}, for both the label-only attack on $S$ and the white-box attacks of Section~\ref{sec:whitebox}, and ROC curves are shown in Figure~\ref{fig:roc_combined}.

\begin{figure}[t]
\centering
\includegraphics[width=0.9\columnwidth]{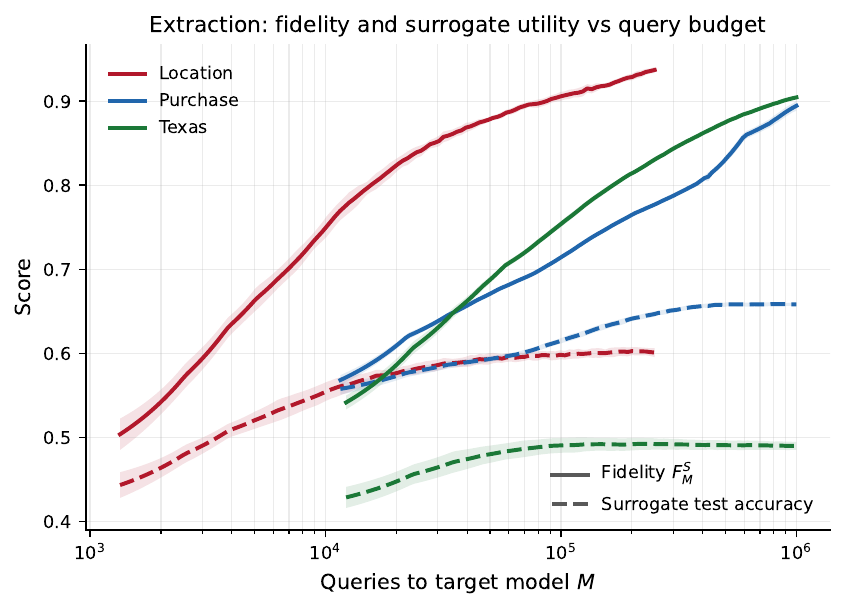}
\caption{Extraction across all three datasets: surrogate fidelity $F_M^S$ (solid) and surrogate test accuracy (dashed) against the number of queries issued to $M$.}
\label{fig:extraction_all}
\end{figure}

Across all datasets, we observe that increasing the query budget leads to consistent and often non-linear improvements in both model fidelity and membership inference performance. Figure~\ref{fig:extraction_all} plots the extraction trajectories side by side. The three fidelity curves converge to a similar band at their largest budgets but reach it at very different rates, with Location requiring roughly an order of magnitude fewer queries than Purchase or Texas. Surrogate test accuracy plateaus at each target's own test accuracy while fidelity keeps rising, so beyond that point additional queries buy agreement with $M$'s decision function rather than better generalization. Since the decision function carries the membership signal, further querying continues to help the attack even though it no longer improves the surrogate as a classifier. On the Location dataset, $S_{1\text{K}}$ achieves a fidelity of $0.4815$ and a label-only AUC of $0.5867$, which rise to $0.9087$ and $0.8546$ at $S_{100\text{K}}$. The ROC curves in Figure~\ref{fig:roc_location} show that $M$ is not fully extracted at $q=100\text{K}$, yet the membership inference accuracies are within $2$--$3\%$ of those obtained against $M$ directly. Increasing $q$ by a further $2.5\times$ to $S_{250\text{K}}$ raises fidelity to $0.9283$ but leaves the label-only AUC essentially unchanged at $0.8567$, indicating that the label-only signal recoverable from the surrogate saturates before fidelity does. The white-box attacks close and then reverse this gap at the same budget, reaching $0.8817$ AUC against the target's $0.8725$. ROC curve of $S_{\text{10K}}$ is closer to its successor $S_{\text{100K}}$ than the ($S_{\text{100K}}$, $S_{\text{1M}}$) pairs of Purchase and Texas datasets. This can be explained with the lesser complexity of $M$ (due to smaller $n_c$, $n_j$ and $|D_M|$) trained on Location dataset.

On the Purchase dataset, the surrogate's fidelity reaches $0.8732$ and label-only AUC reaches $0.7997$ at $S_{1\text{M}}$, against the target model’s AUC of $0.8401$. The ROC curves in Figure~\ref{fig:roc_purchase} highlight that $S_{\text{100K}}$ has the least similarity with $M$ among the three datasets, which is reflected as a sharp decrease in membership inference attack accuracy and AUC. The white-box attack on the same $S_{\text{1M}}$ surrogate reaches $0.8957$ AUC, exceeding the target by $5.6$ percentage points, so the residual label-only gap reflects the restrictiveness of the hard-label interface rather than a loss of membership signal in the surrogate. This allows an unlimited number of samples to be inferred offline and potentially enables more advanced attacks (e.g., model inversion) against $M$. Considering the complexity of $M$ ($n_c$, $n_j$ and $|D_M|$) trained on Purchase dataset, $q=1M$ is sufficient enough for almost full extraction.  

The Texas dataset shows a more gradual trajectory. At $S_{1\text{M}}$, the surrogate achieves $0.8981$ fidelity and $0.8082$ label-only AUC, still below the target’s $0.8363$ AUC, while the white-box attack on the same surrogate reaches $0.8863$. The slower convergence suggests that high-dimensional, sparse, or less structured datasets require more extensive exploration to accurately approximate membership-sensitive regions. Furthermore, comparison between the ROC curves of $S_{\text{100K}}$ in Texas and $S_{\text{10K}}$ in Location demonstrate that similar $F_M^S$ does not necessarily translate into similar attack accuracy or AUC in membership inference. This emphasizes that decision boundary alignment in areas critical to high-precision privacy attacks is more difficult to replicate in complex feature spaces especially under low query budget regimes.

The results demonstrate that high surrogate fidelity and sufficient query budget $q$ are critical for strong membership inference performance, as the models get more complex and the data gets more sparse. On Location, $S_{10\text{K}}$ achieves an AUC of $0.7986$ with only $0.7602$ fidelity, suggesting that partial boundary recovery may suffice in lower-dimensional or well-generalized settings. In contrast, for Purchase and Texas, meaningful replication of membership leakage only emerges once fidelity exceeds roughly $0.87$ and $q> 100K$, indicating that more complex or sparse feature spaces require tighter alignment with the target’s decision boundaries. These findings highlight the potential need for further querying or targeted optimization over underrepresented classes or feature regions.

\begin{table*}[ht]
\centering
\caption{Performance summary of privacy attacks under varying query budgets.
Black-box columns use the label-only boundary-distance attack on the surrogate
$S$. Each white-box column independently reports the best of loss, max, SD,
entropy and SeqMIA at that budget; the \textbf{Attack(s)} column names the
winner of each metric in column order (AUC / Att.\ Acc.\ / TPR@0.1\% /
TPR@1\%), abbreviated to a single name when one attack wins all four. Values
exceeding the corresponding value for the target $M$ are shown in bold. All
entries are mean $\pm$ standard deviation over 10 runs.}
\begin{adjustbox}{max width=\textwidth}
\begin{tabular}{l l cc cccc l cccc}
\toprule
\textbf{Dataset} & \textbf{Budget} & \multicolumn{2}{c}{\textbf{Extraction}} & \multicolumn{4}{c}{\textbf{Black-box MIA on }$\bm{S}$} & \multicolumn{5}{c}{\textbf{White-box MIA on }$\bm{S}$} \\
\cmidrule(lr){3-4} \cmidrule(lr){5-8} \cmidrule(lr){9-13}
 & & $\bm{F_M^S}$ \textbf{(\%)} & \textbf{Test Acc.\ (\%)} & \textbf{AUC (\%)} & \textbf{Att.\ Acc.\ (\%)} & \textbf{TPR@0.1\% (\%)} & \textbf{TPR@1\% (\%)} & \textbf{Attack(s)} & \textbf{AUC (\%)} & \textbf{Att.\ Acc.\ (\%)} & \textbf{TPR@0.1\% (\%)} & \textbf{TPR@1\% (\%)} \\
\midrule
\textbf{Location} & $S_{1K}$ & $48.15 \pm 1.22$ & $42.79 \pm 1.10$ & $58.67 \pm 1.29$ & $58.80 \pm 1.23$ & $0.10 \pm 0.08$ & $0.94 \pm 0.36$ & Loss/Loss/Seq/Loss & $61.07 \pm 1.36$ & $59.57 \pm 0.82$ & $0.13 \pm 0.12$ & $0.75 \pm 0.20$ \\
 & $S_{10K}$ & $76.02 \pm 1.19$ & $56.00 \pm 0.87$ & $79.86 \pm 1.00$ & $76.57 \pm 1.00$ & $0.24 \pm 0.25$ & $\bm{2.19 \pm 0.94}$ & Loss/Loss/Seq/Seq & $81.58 \pm 1.03$ & $77.52 \pm 0.82$ & $\bm{0.68 \pm 0.49}$ & $\bm{3.54 \pm 1.42}$ \\
 & $S_{100K}$ & $90.87 \pm 0.55$ & $59.75 \pm 0.62$ & $85.46 \pm 1.19$ & $84.16 \pm 0.92$ & $0.27 \pm 0.25$ & $1.90 \pm 0.63$ & Loss/Loss/Seq/Seq & $\bm{87.88 \pm 0.83}$ & $86.47 \pm 0.77$ & $\bm{1.07 \pm 0.65}$ & $\bm{5.31 \pm 1.99}$ \\
 & $S_{250K}$ & $92.83 \pm 0.41$ & $59.78 \pm 0.52$ & $85.67 \pm 1.01$ & $84.82 \pm 0.69$ & $\bm{0.34 \pm 0.30}$ & $1.98 \pm 0.86$ & Loss/Loss/Seq/Seq & $\bm{88.17 \pm 0.85}$ & $\bm{87.50 \pm 0.52}$ & $\bm{1.03 \pm 0.87}$ & $\bm{5.61 \pm 2.65}$ \\
 & $M$ (target) & --- & $60.16 \pm 0.47$ & $87.25 \pm 1.13$ & $86.98 \pm 0.83$ & $0.29 \pm 0.23$ & $1.99 \pm 0.95$ & \multicolumn{5}{c}{\textit{not applicable: $M$ returns hard labels only}} \\
\midrule
\textbf{Purchase} & $S_{10K}$ & $56.61 \pm 0.45$ & $53.55 \pm 0.48$ & $57.66 \pm 0.81$ & $57.24 \pm 0.82$ & $0.12 \pm 0.10$ & $1.23 \pm 0.53$ & Loss & $59.23 \pm 0.87$ & $57.63 \pm 0.63$ & $\bm{0.19 \pm 0.24}$ & $1.23 \pm 0.43$ \\
 & $S_{100K}$ & $73.07 \pm 0.23$ & $62.86 \pm 0.42$ & $69.52 \pm 1.07$ & $67.13 \pm 0.87$ & $\bm{0.23 \pm 0.28}$ & $1.41 \pm 0.52$ & SeqMIA & $71.71 \pm 1.08$ & $67.66 \pm 1.04$ & $\bm{0.39 \pm 0.52}$ & $\bm{2.90 \pm 0.80}$ \\
 & $S_{500K}$ & $81.38 \pm 0.34$ & $\bm{65.45 \pm 0.26}$ & $75.34 \pm 1.17$ & $72.95 \pm 0.75$ & $0.10 \pm 0.11$ & $1.48 \pm 0.39$ & SeqMIA & $81.05 \pm 1.11$ & $75.18 \pm 0.99$ & $\bm{1.02 \pm 0.93}$ & $\bm{6.42 \pm 1.13}$ \\
 & $S_{1M}$ & $87.32 \pm 0.67$ & $\bm{65.51 \pm 0.20}$ & $79.97 \pm 0.95$ & $78.20 \pm 0.75$ & $0.11 \pm 0.13$ & $1.60 \pm 0.46$ & SeqMIA & $\bm{89.57 \pm 0.75}$ & $\bm{83.97 \pm 0.50}$ & $\bm{1.01 \pm 0.89}$ & $\bm{9.48 \pm 2.59}$ \\
 & $M$ (target) & --- & $65.04 \pm 0.20$ & $84.01 \pm 0.85$ & $83.31 \pm 0.68$ & $0.13 \pm 0.13$ & $1.80 \pm 0.44$ & \multicolumn{5}{c}{\textit{not applicable: $M$ returns hard labels only}} \\
\midrule
\textbf{Texas} & $S_{10K}$ & $53.47 \pm 0.71$ & $42.26 \pm 1.13$ & $54.51 \pm 0.61$ & $55.15 \pm 0.53$ & $0.11 \pm 0.10$ & $0.99 \pm 0.38$ & Loss/Loss/Ent/Loss & $56.51 \pm 0.70$ & $56.73 \pm 0.66$ & $0.19 \pm 0.14$ & $0.70 \pm 0.26$ \\
 & $S_{100K}$ & $76.38 \pm 0.37$ & $\bm{48.84 \pm 0.79}$ & $72.28 \pm 0.92$ & $71.52 \pm 0.81$ & $0.16 \pm 0.16$ & $1.72 \pm 0.57$ & SeqMIA & $77.92 \pm 0.90$ & $72.06 \pm 0.53$ & $\bm{1.59 \pm 1.03}$ & $\bm{6.30 \pm 1.41}$ \\
 & $S_{500K}$ & $87.68 \pm 0.18$ & $\bm{48.84 \pm 0.56}$ & $80.75 \pm 0.60$ & $79.32 \pm 0.77$ & $0.30 \pm 0.19$ & $2.06 \pm 0.75$ & SeqMIA & $\bm{87.93 \pm 0.66}$ & $80.99 \pm 0.73$ & $\bm{3.69 \pm 1.60}$ & $\bm{13.69 \pm 3.34}$ \\
 & $S_{1M}$ & $89.81 \pm 0.20$ & $\bm{49.00 \pm 0.47}$ & $80.82 \pm 0.69$ & $79.89 \pm 0.67$ & $0.27 \pm 0.32$ & $1.98 \pm 0.64$ & SeqMIA & $\bm{88.63 \pm 0.75}$ & $82.07 \pm 0.82$ & $\bm{3.53 \pm 1.93}$ & $\bm{15.10 \pm 3.23}$ \\
 & $M$ (target) & --- & $48.37 \pm 0.29$ & $83.63 \pm 0.70$ & $82.71 \pm 0.59$ & $0.52 \pm 0.49$ & $2.38 \pm 0.91$ & \multicolumn{5}{c}{\textit{not applicable: $M$ returns hard labels only}} \\
\bottomrule
\end{tabular}
\end{adjustbox}
\label{tab:query_budget_summary}
\end{table*}

\textbf{White-box attacks on the surrogate.} The label-only results above restrict the attacker to the same hard-label interface on $S$ that $M$ imposes, which isolates the effect of amortization. Removing that restriction changes the picture. As Table~\ref{tab:query_budget_summary} shows, at the largest budget the strongest white-box attack on $S$ exceeds the label-only attack run directly on $M$ in AUC on all three datasets, by $0.9$, $5.6$ and $5.0$ percentage points on Location, Purchase and Texas respectively, while matching its balanced attack accuracy to within $0.7$ percentage points. The margin is wider in the low false-positive regime: TPR@$1\%$FPR rises from $1.99\%$ to $5.61\%$ on Location, from $1.80\%$ to $9.48\%$ on Purchase, and from $2.38\%$ to $15.10\%$ on Texas, improvements of $2.8\times$, $5.3\times$ and $6.3\times$. In auditing terms, on Texas an adversary willing to accept one false accusation per hundred non-members identifies roughly one member in seven rather than one in forty. Since the target returns only hard labels, none of these attacks can be mounted against $M$ at all; the surrogate is what makes them available.

The gain is not uniform across attacks, and the two metric families do not always rank them the same way. On Purchase and Texas, SeqMIA leads on every metric once fidelity exceeds roughly $0.73$, which is consistent with its use of the full distillation trajectory rather than the surrogate's final state. On Location the two disagree: a simple loss threshold attains the highest mean AUC at every budget above $1\text{K}$, but its separation in the low-FPR region collapses to $0.00\%$ TPR@$1\%$FPR at $S_{250\text{K}}$, while SeqMIA gives up under two points of AUC and returns $5.61\%$. This matches the observation of Carlini et al.~\cite{carlini_lira} that average-case and low-FPR behavior are largely uncorrelated. The choice of attack on the surrogate should therefore follow the auditing objective rather than a single headline number, and we report both metrics in Table~\ref{tab:query_budget_summary} together with the winning attack for each.

\subsection{Comparison with Recent Label-Only Attacks}
\label{sec:comparison}

A direct head-to-head with recent label-only attacks is not possible: they occupy a different point in the assumption space, and their tabular configurations differ in training set size and hence in overfitting level. We therefore report the comparison as indicative and state the discrepancies explicitly.

On Location, YOQO~\cite{yoqo} reports membership inference accuracies of $77.50\%$ (online) and $83.25\%$ (offline), against $82.27\%$ for the boundary attack and $70.94\%$ for the gap attack, on a target trained with $|D_M| = 1{,}000$. Our white-box attack on $S_{250\text{K}}$ attains $87.50\%$ attack accuracy with $|D_M| = 1{,}600$. On Purchase, YOQO reports $70.00\%$ (offline) and $68.28\%$ (online) against $74.17\%$ for the boundary attack at $|D_M| = 6{,}000$, where we attain $83.97\%$ at $|D_M| = 10{,}000$. DHAttack~\cite{enhanced-lomia} reports on Purchase an AUC of $0.727$ and TPR@$0.1\%$FPR of $0.34\%$ at $|D_M| = 20{,}000$, against the strongest baseline in their evaluation at $0.738$ AUC and $0.21\%$; our white-box attack on $S_{1\text{M}}$ attains $0.8957$ AUC and $1.01\%$ TPR@$0.1\%$FPR. OSLO~\cite{oslo} evaluates only on image benchmarks and reports no tabular results, so no comparison is available there. Because larger training sets reduce overfitting and therefore membership leakage, the differing $|D_M|$ favors our numbers on Purchase and disfavors them on Location; we do not claim these gaps are attributable to the attacks alone.

The larger difference is in what each attacker is assumed to hold. YOQO trains $16$ in-out model pairs, and its online variant retrains the in-models for every target sample; OSLO trains between $6$ and $18$ surrogate models; DHAttack trains up to $256$ shadow models and further assumes the target's architecture and hyperparameters are known. All three require these models to be trained on data drawn from the target's training distribution, an assumption OSLO explicitly identifies as a limitation of its design. DHAttack additionally relabels a reference set of twice the training set size through the target, a one-time cost of the same order as our entire extraction budget on Location. Our attacker assumes none of this: it observes hard labels, holds no distributional or architectural knowledge, trains no shadow model aligned with $D_M$, and requires at most a small seed that Section~\ref{sec:da_sensitivity} shows can be removed entirely at a modest cost in convergence speed. These two lines of work are complementary. Their contribution is a stronger per-sample primitive under a more informed adversary; ours is that extraction both amortizes the per-sample cost and removes the interface restriction that defines the label-only setting. Since our surrogate is an ordinary white-box model, these attacks can also be applied to it, which we leave as an extension.

\subsection{Auxiliary Seed Sensitivity}
\label{sec:da_sensitivity}
We vary the size of seed set $D_A$ on the Location dataset to characterize its role from a near data-free setting ($|D_A|=0$) up to $|D_A|=300$, and also evaluate a distribution-shifted seed drawn from a class-imbalanced subpopulation. Appendix~\ref{app:seed} reports the full results. In summary, at $q=100\text{K}$ the attack is essentially insensitive to the seed: fidelity stays near $0.90$ and AUC near $0.84$ across the entire range, and even the data-free setting reaches $F_M^S=0.82$ and AUC $=0.81$. At $q=10\text{K}$ the seed matters, but as a convergence accelerant rather than a ceiling, and the distribution-shifted seed behaves the same way. Both a smaller seed and a mismatched seed slow convergence rather than cap performance, and additional querying absorbs either handicap. This supports our characterization of $D_A$ as a bootstrap rather than a standing requirement (Section~\ref{sec:systhreatmodel}).

\subsection{Cost-Effectiveness}
Label-only membership inference attacks such as the one proposed by~\cite{choquette} achieve strong MIA performance, but they incur significant query costs. In their reported experiments, they achieve membership accuracies of $0.8920$, $0.8740$, and $0.8030$ on Location, Purchase, and Texas, respectively. Since extraction is a one-time cost while direct label-only inference pays $c$ queries per victim sample, amortization becomes favorable once the number of audited samples exceeds $n^\star = q/c$, where $q$ is the budget required to extract a surrogate of comparable attack quality. We measure $c$ from our own implementation rather than adopting the figure reported by prior work: the boundary-distance attack issues $2{,}388$, $2{,}365$ and $2{,}398$ queries per sample on Location, Purchase and Texas. These are well below the $10{,}000$ commonly quoted, so adopting them makes our break-even analysis conservative. Table~\ref{tab:breakeven} reports $n^\star$ for each dataset, paired with the surrogate and target AUC at that budget. A high-quality surrogate breaks even after roughly $105$ audited samples on Location and $420$ on Purchase and Texas. Below these counts, direct label-only inference remains cheaper; above them, the offline surrogate is reused at no further query cost, and the advantage grows without bound as the audited population increases.

\begin{table}[ht]
\centering
\caption{Break-even count $n^\star = q/c$ at which extraction-based auditing becomes cheaper than direct label-only MIA. $c$ is the measured mean queries to $M$ per audited sample at \texttt{max\_iter}~$=10$. Values exceeding the target are bold.}
\begin{adjustbox}{width=\columnwidth}
\begin{tabular}{l c r r r ccc}
\toprule
\textbf{Dataset} & \textbf{Surrogate} & $\mathbf{q}$ & $\mathbf{c}$ & $\mathbf{n^\star}$ & \multicolumn{3}{c}{\textbf{AUC}} \\
\cmidrule(lr){6-8}
 & & & & & $\bm{S_{BlackBox}}$ & $\bm{S_{WhiteBox}}$ & $\bm{M}$ \\
\midrule
Location & $S_{250\text{K}}$ & 250{,}000     & 2{,}388 & 105 & 0.8567 & $\bm{0.8817}$ & 0.8725 \\
Purchase & $S_{1\text{M}}$   & 1{,}000{,}000 & 2{,}365 & 423 & 0.7997 & $\bm{0.8957}$ & 0.8401 \\
Texas    & $S_{1\text{M}}$   & 1{,}000{,}000 & 2{,}398 & 417 & 0.8082 & $\bm{0.8863}$ & 0.8363 \\
\bottomrule
\end{tabular}
\end{adjustbox}
\label{tab:breakeven}
\end{table}

Our method may not always be the preferable option: if the number of samples to be evaluated is small, direct label-only membership inference, despite its high per-sample cost, may remain the more efficient route. Our method is more appropriate for the attackers seeking to audit or infer membership for large populations, where the one-time cost of surrogate extraction can be justified. The surrogates not only approximate the prediction function of the target model, but also reproduce its decision boundary vulnerabilities, highlighting a viable and generalizable pathway for privacy leakage in label-only black-box settings. Past break-even the advantage grows linearly with the audited population, since the extracted surrogate is reused at no further query cost while direct inference keeps paying $c$ per sample. The amortized cost per audited sample therefore tends to zero as the population grows, which is the large-scale auditing regime our attack targets. This also changes the shape of the interaction with $M$: amortization converts a continuous, detectable querying load into a one-time, lower-profile one, which is as much an operational advantage as a numerical one. 

\section{Countermeasures}
\label{sec:countermeasures}

\begin{table*}[ht]
\centering
\caption{Effect of countermeasures on attack and model performance in the Location dataset.}
\begin{adjustbox}{max width=\textwidth}
\begin{tabular}{ccccccccccc}
\toprule
\multirow{2}{*}{\textbf{Defensive Strategy}} & \multirow{2}{*}{\textbf{Setting}} 
& \multicolumn{4}{c}{$M'$ Metrics}
& \multicolumn{5}{c}{$S$ Metrics}  \\
\cmidrule(r){3-6}
\cmidrule(r){7-11}
& & $F_M^{M'}$ & Test Acc (\%) & Attack Acc (\%) & \textbf{AUC} & $F_M^S$ & $F_{M'}^S$ & Test Acc (\%) & Attack Acc (\%) & \textbf{AUC} \\
\midrule

\multirow{4}{*}{\textbf{DP-SGD}} 
& $\epsilon=20$ & 0.6598 & 59.96 & 73.67 & \textbf{0.7636} & 0.6621 & 0.8541 & 60.02 & 72.00 & \textbf{0.7351} \\
& $\epsilon=50$ & 0.7044 & 60.05 & 82.33 & \textbf{0.8383} & 0.7021 & 0.8605 & 61.21 & 79.00 & \textbf{0.8178} \\
& $\epsilon=100$ & 0.7060 & 61.16 & 84.33 & \textbf{0.8488} & 0.7076 & 0.8612 & 60.76 & 78.00 & \textbf{0.7916} \\
& $\epsilon=200$ & 0.7169 & 60.87 & 84.00 & \textbf{0.8327} & 0.7012 & 0.8500 & 59.99 & 81.33 & \textbf{0.8305} \\
\midrule

\multirow{4}{*}{\textbf{Dropout}} 
& $p=0.2$ & 0.8682 & 61.89 & 87.67 & \textbf{0.8546} & 0.8480 & 0.9080 & 61.40 & 85.67 & \textbf{0.8442} \\
& $p=0.4$ & 0.8548 & 60.28 & 88.00 & \textbf{0.8674} & 0.8391 & 0.9077 & 60.40 & 84.00 & \textbf{0.8636} \\
& $p=0.6$ & 0.8423 & 60.13 & 87.00 & \textbf{0.8879} & 0.8275 & 0.9157 & 59.15 & 83.00 & \textbf{0.8448} \\
& $p=0.8$ & 0.8028 & 61.28 & 84.67 & \textbf{0.8500} & 0.7932 & 0.9272 & 61.43 & 82.00 & \textbf{0.8415} \\
\midrule

\multirow{4}{*}{\textbf{L2 Reg.}} 
& $\lambda = 0.0001$  & 0.8535 & 62.91 & 86.67 & \textbf{0.8581} & 0.8346 & 0.9160 & 61.40 & 85.67 & \textbf{0.8548} \\
& $\lambda = 0.0005$  & 0.8378 & 63.23 & 86.67 & \textbf{0.8532} & 0.8195 & 0.9182 & 62.36 & 84.33 & \textbf{0.8461} \\
& $\lambda = 0.001$  & 0.8269 & 63.79 & 84.67 & \textbf{0.8572} & 0.8086 & 0.9186 & 63.26 & 84.00 & \textbf{0.8614} \\
& $\lambda = 0.005$  & 0.8102 & 66.60 & 81.67 & \textbf{0.8312} & 0.7967 & 0.9160 & 66.40 & 82.00 & \textbf{0.8300} \\
\midrule

\multirow{4}{*}{\textbf{Undefended (M)}} &  
& \multicolumn{4}{c}{$M$ Metrics}
& \multicolumn{5}{c}{$S$ Metrics}  \\
\cmidrule(r){3-6}
\cmidrule(r){7-11}
& & $F_M^{M'}$ & Test Acc (\%) & Attack Acc (\%) & \textbf{AUC} & $F_M^S$ & $F_{M'}^S$ & Test Acc (\%) & Attack Acc (\%) & \textbf{AUC} \\
\cmidrule(r){3-11}
& & -- & 60.78 & 88.67 & \textbf{0.8919} & 0.9128 & \textbf{--} & 59.99 & 88.33 & \textbf{0.8964} \\
\bottomrule
\end{tabular}
\end{adjustbox}
\label{tab:defense_strategy_metrics}
\end{table*}

We evaluate the impact of three training-time defenses, DP-SGD~\cite{dpsgd}, Dropout~\cite{dropout}, and L2 regularization~\cite{andrewng, l2}, on the performance of our attack. These act on the model rather than on its outputs, which is deliberate: because our attack is label-only and runs through an extracted surrogate, defenses that perturb or mask confidence scores (e.g., MemGuard) leave it untouched, the same limitation Li and Zhang~\cite{li_and_zhang} report for decision-based attacks. The three we consider instead target the generalization gap and the decision boundary that the attack exploits, and all have reported effectiveness against label-only membership inference~\cite{choquette}. All models are trained on the same training split of the Location dataset, with only the applied defense varying. We denote the defended target as $M'$ and the undefended target as $M$, and summarize the results in Table~\ref{tab:defense_strategy_metrics}.

The undefended model $M$ is the most vulnerable, yielding AUC scores of $0.8919$ and $0.8964$ when attacked directly and through its surrogate $S$, respectively. The near-identical values confirm substantial membership leakage in both the target and the surrogate, so in the absence of any defense the attack remains highly effective end to end. Among the evaluated defenses, DP-SGD reduces attack success the most. At $\epsilon=20$, where the noise injection is largest, AUC falls to $0.7636$ on $M'$ and $0.7351$ on $S$, well below the undefended case, and a lower $\epsilon$ would reduce it further. This protection, however, comes from degrading the model itself: the defended target diverges from the original, as the low $F_M^{M'}$ scores show, so the privacy gain is inseparable from a loss of fidelity and utility. We also note that even at this setting the surrogate AUC stays near $0.73$, well above chance, so DP-SGD curtails the attack without eliminating it.

Dropout is less disruptive to utility but also less effective. Even at $p=0.8$, AUC on $M'$ and $S$ remains elevated and the surrogate continues to track the target closely, so dropout alone does little to hinder either the extraction or the inference stage. L2 regularization behaves similarly: increasing $\lambda$ lowers extraction fidelity and modestly reduces inference success while improving the generalization accuracy of $M'$, but at the strongest setting ($\lambda=0.005$) AUC drops by at most $6$--$7\%$ relative to the undefended case.

Overall, DP-SGD provides the most substantial privacy gains, though at a clear utility cost, while L2 regularization is a reasonable complement to it because it raises the test accuracy of $M'$, and dropout offers the least of the three. Across all settings the surrogate remains a faithful and reusable stand-in for the defended target, which indicates that training-time defenses constrain but do not close the label-only leakage channel; query-side defenses against the extraction stage itself, such as query monitoring or rate-limiting, are complementary and orthogonal to those studied here.

We evaluate these defenses against the label-only attack path, and this choice is deliberate rather than incidental. The label-only attack on $S$ is the weakest version of our framework: it discards the posteriors and the training trajectory that the attacker holds, and Section~\ref{sec:eval} shows that the white-box attacks outperform it at every budget we examined. Its performance under a defense is therefore a lower bound on what a defended deployment faces, and the bound remaining high is the informative result. Dropout and $\ell_2$ regularization leave surrogate AUC between $0.83$ and $0.86$ against $0.8964$ undefended, a reduction of at most about seven percent, and DP-SGD reaches $0.7351$ only by degrading the target itself. Since none of the three brings even this weakest attack near chance, subjecting the better-informed white-box attacks to the same defenses could only widen the reported gap, never close it. Measuring them would strengthen our claims rather than qualify them, so we report the conservative case. This also indicates that defenses acting on the training procedure address the wrong stage of the attack. What they suppress is the membership signal in $M$, while what makes the surrogate useful to an attacker is that it reproduces $M$'s decision geometry at all. Defenses that target the extraction stage, such as query monitoring, rate limiting or detection of adaptive sampling, act on the mechanism our framework depends on, and are the more promising direction.

\section{Extending Label-Only Membership Inference to Deep Neural Networks}
\label{sec:dnn_extension}

Our primary evaluation focuses on models trained on tabular data. However, the proposed framework is not inherently restricted to this setting. In this section, we investigate the extent to which the same surrogate-based, label-only membership inference pipeline can be instantiated for deep neural networks trained on image data using a different kind of membership inference attack. Our objective is not to establish state-of-the-art attack performance on image benchmarks, but rather to assess the feasibility, limitations, and qualitative behavior of extraction-based membership inference when applied to high-capacity models operating on high-dimensional inputs, under the same strict black-box constraints considered throughout the paper.

The experiments in this section are preliminary. They are intended to probe how far the core design principles of our framework, namely query-efficient surrogate extraction followed by offline membership inference, extend beyond the tabular domain, and to locate where they break down. Our intent is twofold. First, we establish a well-defined baseline for membership inference on modern image classifiers using SeqMIA~\cite{seqmia}, which constructs membership signals from sequential model behavior during distillation. Second, we integrate an adaptive-sampling-based model extraction stage from out-of-distribution (OOD) datasets under the same label-only constraints and evaluate the extent to which transferring query overhead into a one-time surrogate construction phase remains viable in high-dimensional settings.

\subsection{Threat Model and Experimental Protocol}
\label{subsec:dnn_protocol}

We adopt the same strict label-only black-box threat model used in our tabular experiments. The attacker may query a trained image classifier $M$ and observes only the predicted class label. We do not assume access to confidence scores, intermediate activations, gradients, architectural details, training data, or the training data distribution. The only difference from the tabular setting is that, for deep learning, the attacker may leverage out-of-distribution (OOD) data sources, provided that these do not overlap with the training data of $M$. This is motivated by the practical difficulty of generating diverse and meaningful candidate queries in pixel space without using generative models or without spending additional queries to synthesize informative samples, which conflicts with the low-query objective of our setting. In particular, we do not assume the availability of surrogate models trained on data drawn from the same distribution as the target model. This constraint distinguishes our setting from prior work~\cite{enhanced-lomia, yoqo, oslo, seqmia}, which often relies on auxiliary datasets closely aligned with the target distribution or access to soft prediction outputs.

We conduct all image experiments on CIFAR-10~\cite{cifar}. Following the SeqMIA protocol, we repartition it into three disjoint subsets of 20{,}000 samples for target training/testing, surrogate training/testing, and distillation, so that target and surrogate are trained on disjoint but in-distribution data. Appendix~\ref{app:image_protocol} gives the exact splits. All results in this section are reported under this fixed split regime. Importantly, our goal is not to induce overfitting as a means to amplify membership leakage. Instead, we intentionally train high-utility target and surrogate models with strong test accuracy to more closely reflect operational regimes typical of MLaaS deployments, where standard training pipelines employ augmentation and regularization.

\subsection{Methodology for Baseline and Sampling-Assisted Membership Inference}
\label{subsec:dnn_methodology}

We evaluate two standard convolutional architectures for target and surrogate models' training: ResNet-50~\cite{resnet} and VGG-16~\cite{vgg16}. For each architecture, we train a target model $M$ on the CIFAR-10 target training split and an ID surrogate model on the CIFAR-10 surrogate training split, using the same hyperparameter configuration for a given architecture. Both architectures are trained with SGD and data augmentation for 100 epochs; the per-architecture schedules are given in Appendix~\ref{app:image_protocol}. The same recipe is applied to the target and to the ID surrogate.

\textbf{SeqMIA baseline:} The attack is the one described in Section~\ref{sec:whitebox}; here we give only the settings specific to the image models. We distill each model for 50 epochs and record the distilled checkpoint at every epoch from 1 through 50. The attack model is a single-layer LSTM with 50 hidden units and an attention mechanism, trained for 150 epochs using Adam with learning rate $5\times 10^{-4}$. We employ StepLR with step size 50 and $\gamma=0.1$, reducing the learning rate by a factor of 10 at epochs 50 and 100.

\textbf{Adaptive-sampling-assisted variant with an out-of-distribution (OOD) query pool:} We incorporate a query-efficient extraction stage to negate the reliance of SeqMIA on ID auxiliary training data. We perform a model extraction attack against the same trained target model $M$ used in the baseline by using adaptive-sampling under label-only black-box access. Unlike our tabular instantiation, where candidates are synthesized directly through feature-space perturbations, pool-free synthesis is impractical in pixel space. Here we therefore draw adaptive query candidates from STL-10~\cite{stl10}, a public OOD image dataset. We initialize extraction with a seed set of 500 CIFAR-10 samples that are pre-labeled by the target model. Extraction proceeds for 100 iterations. In each iteration, candidate images from STL-10 are selected, queried to $M$ for labels, and used to update the surrogate. The surrogate is trained for a single epoch per iteration using SGD (learning rate $10^{-1}$, momentum $0.9$, weight decay $10^{-7}$, batch size 128), and StepLR (step size 5, $\gamma=0.9$) is used as the learning rate scheduler. We set the total query budget to 10{,}000, which matches the size of the CIFAR-10 target training split. We set $\gamma_1=\gamma_2=0.7$, so that approximately half of the candidate pool is filtered at each stage, consistent with our prior instantiation.

We reuse the final query-labeled set to train an alternative surrogate model from scratch as an additional robustness measure. This step addresses a practical concern: if a surrogate model trained on OOD queries has low in-distribution utility, it may not reflect membership-relevant behavior of the target model. Training a fresh surrogate model on the target model labeled set provides a controlled way to evaluate how much membership information can be recovered when the surrogate training data is derived from OOD queries rather than from the target distribution. We refer to this model as the OOD surrogate model.

\subsection{Results and Limitations}
\label{subsec:dnn_results}

Table~\ref{tab:image_results} summarizes the test accuracy and membership inference performance for both architectures. We report the test accuracy of the target model, the ID surrogate model trained under the standard SeqMIA split, and the OOD surrogate model trained on the target model labeled query set. We then report membership inference AUC and membership prediction accuracy for (i) the SeqMIA baseline and (ii) the OOD-surrogate variant, where the ID surrogate model is replaced by the OOD surrogate model trained from label-only queries.

\begin{table}[t]
\centering
\caption{CIFAR-10 test accuracies and membership inference performance. ``ID'' is the standard SeqMIA surrogate; ``OOD'' is ours, trained from label-only STL-10 queries. Membership results are over 20k samples. Top results report the test accuracies of models and the bottom results report the membership inference accuracies as percentages and AUCs.}
\label{tab:image_results}
\begin{adjustbox}{width=\columnwidth}
\begin{tabular}{lcccc}
\toprule
\multicolumn{1}{l}{\textbf{Models}} & \multicolumn{2}{c}{\textbf{ResNet-50 Acc.(\%)}} & \multicolumn{2}{c}{\textbf{VGG-16 Acc.(\%)}} \\
\cmidrule(lr){2-3}\cmidrule(lr){4-5}
Target & \multicolumn{2}{c}{90.84} & \multicolumn{2}{c}{82.66} \\
ID Surrogate & \multicolumn{2}{c}{90.12} & \multicolumn{2}{c}{81.61} \\
OOD Surrogate & \multicolumn{2}{c}{63.77} & \multicolumn{2}{c}{62.93} \\
\midrule
\multicolumn{1}{l}{\textbf{Models}} & \textbf{AUC} & \textbf{Acc.(\%)} & \textbf{AUC} & \textbf{Acc.(\%)} \\
ID Surrogate $\approx M$ & 0.5769 & 51.16 & 0.6310 & 58.14 \\
OOD Surrogate & 0.5621 & 50.03 & 0.5615 & 51.67 \\
\bottomrule
\end{tabular}    
\end{adjustbox}
\end{table}

The target models achieve high utility by construction, and under these regimes the ID baseline membership inference performance is modest (Table~\ref{tab:image_results}). This behavior is expected in high-utility settings, where augmentation and regularization reduce overfitting and therefore attenuate the membership signals exploited by MIAs.

The OOD surrogate models trained from label-only STL-10 queries lose roughly 26 percentage points of CIFAR-10 test accuracy relative to the ID surrogates, which is expected under our threat model: the surrogate is learned exclusively from label-only feedback, using a fixed budget of 10{,}000 queries and samples drawn from a distribution that differs from CIFAR-10. SeqMIA relies on sequential metrics that are sensitive to the training trajectory of the surrogate. Under these constraints, recovering a decision rule that fully mirrors the target’s in-distribution behavior is inherently challenging.

The more informative comparison is in AUC, the threshold-independent metric we rely on throughout the paper. The OOD surrogate attains 0.5621 AUC on ResNet-50 and 0.5615 on VGG-16. For ResNet-50 this nearly matches the ID baseline of 0.5769, and for VGG-16 it recovers a substantial, though smaller, fraction of the 0.6310 ID baseline. The significance is that the OOD surrogate approaches the ID baseline while removing its central assumption, access to data from the target's own distribution, and relying only on label-only queries against an OOD pool. The membership accuracies sit near chance (50.03\% and 51.67\%), but this follows from the modest absolute AUC of these deliberately high-utility, low-overfitting targets, for which even the ID baseline reaches only 0.5769 and 0.6310; at low AUC the best thresholded accuracy is necessarily close to chance, which is why we read the signal from AUC rather than accuracy. The structural advantage of the extraction-first design also holds here: a single surrogate supports offline inference over an unlimited number of candidates, whereas direct label-only inference would exhaust its budget on a handful of samples. Because the remaining gap to the baseline stems from distribution shift and a tight 10{,}000-query budget rather than from the inference mechanism, we expect a larger budget or a closer query pool to push AUC toward the baseline, the same trend our tabular results show as queries grow.

However, the current CIFAR-10 results highlight a central limitation of extraction-assisted membership inference for deep neural networks: if the extracted surrogate or OOD-derived surrogate model does not generalize as well as ID surrogates, it may fail to preserve membership-relevant behavior of the target model. Improving surrogate alignment with the target's decision boundary and training dynamics, without increasing attacker assumptions, is our immediate direction for future work suggested by these findings. We plan to explore additional datasets and architectures, cross-architecture surrogates, sensitivity to the extraction and inference hyperparameters, and countermeasures in the deep learning setting.

\section{Conclusion}
\label{sec:conclusion}

In this paper, we presented a label-only extraction-based membership inference framework that first extracts a reusable surrogate model and then performs membership inference offline on the surrogate. By shifting the query cost from repeated per-sample boundary probing to a one-time extraction phase, the attack amortizes the cost of label-only MIA across large candidate populations. 

On standard tabular benchmarks, the extracted surrogates preserve enough decision-boundary behavior to recover membership signals close to those obtained by attacking the target directly, while requiring substantially fewer live queries once many samples are audited. Because the attacker holds the surrogate, extraction also removes the interface restriction that defines the label-only setting. Posterior- and trajectory-based attacks that cannot be run against a hard-label target become available on $S$, and applying them exceeds the direct label-only inference attacks on $M$ in AUC on all three benchmarks, and by a larger margin at low false positive rates. 

Our break-even analysis shows that extraction-assisted inference becomes preferable after only tens to hundreds of audited samples for membership, depending on the dataset. We also found that common training-time defenses reduce but do not eliminate leakage, and that stronger privacy mechanisms such as DP-SGD provide the most meaningful reduction at the cost of altering the learned decision function. 

Finally, our preliminary CIFAR-10 experiments suggest that the same principle can extend to deep neural networks, but that surrogate alignment under distribution shift remains a central challenge. Future work should improve extraction under stricter data-free or OOD-only conditions, evaluate larger image datasets and architectures, and study defenses that jointly limit extraction and membership leakage.

\appendix

\section*{Ethical Considerations}
\label{sec:ethics}
This work studies membership inference and model extraction solely to understand and mitigate privacy risks in machine learning systems. Our experiments were conducted on public benchmark datasets and locally trained models, and no attempt was made to target real users or deployed services. The proposed attack pipeline is intended as a measurement tool to quantify worst-case leakage under strict label-only access, thereby helping practitioners evaluate whether their models expose unacceptable privacy signals. We recognize that similar techniques could be misused to infer sensitive information about individuals; for this reason, we emphasize query-efficient evaluation, disclosure of limitations, and assessment of standard defenses such as regularization and differential privacy. We do not release any data, models, or code that enable attacks against specific services, and we advocate that ML providers treat label-only APIs as potentially sensitive interfaces. By highlighting the residual risk that remains even when confidence scores are hidden, our goal is to support the design of safer training pipelines, auditing practices, and access controls rather than to facilitate exploitation.

\bibliographystyle{plainurl}
\bibliography{bibliography}

\section{Notation}
\label{app:notation}

Table~\ref{tab:symbols} summarizes the symbols and notation used throughout the paper.

\begin{table}[ht]
\centering
\caption{Commonly Used Symbols and Notations}
\begin{tabularx}{\columnwidth}{|c|X|}
\hline
\textbf{Symbol} & \textbf{Description} \\
\hline
$\mathbf{M}$ & Target (victim) model \\
$\mathbf{S}$ & Surrogate (a.k.a extracted or stolen) model \\
$\mathbf{D_M}$ & Dataset used to train the target model \\
$\mathbf{D_S}$ & Dataset used to train the surrogate model \\
$\mathbf{D_A}$ & Small seed set used to initialize $S^0$ and bootstrap query synthesis \\
$\mathbf{D_{Q}}$ & Augmented and perturbed query dataset used in active learning for training $S$ \\
$\mathbf{D_N}$ & Neutral dataset used for evaluating model fidelity \\
$\mathbf{D_{\text{mem}}}$ & Membership dataset used for evaluating membership inference attack performance \\
$x_i$ & Input sample to the model \\
$y_i$ & Predicted class (label) for the sample $x_i$ \\
$n_j$ & Number of features in the input \\
$n_c$ & Number of output classes \\
$R_j$ & Range of possible values for feature $j$ \\
$R_c$ & Range of possible values for classes (labels) \\
$\tau$ & Decision boundary distance threshold \\
$\mathcal{L}$ & Loss function \\
$\nabla_{\mathbf{\theta}} \mathcal{L}$ & Gradient of the loss with respect to model parameters \\
$F_{\text{M}}^{\text{S}}$ & Fidelity of the surrogate model $S$ relative to the target model $M$ \\
$q$ & Query budget for the attacker \\
\hline
\end{tabularx}
\label{tab:symbols}
\end{table}

\section{Variability and Computational Settings}
\label{app:var_comp}

In the regular and countermeasure experiments, we ran each experiment multiple times (10+) to ensure result stability and to report means and standard deviations of key metrics given different query budgets $q$. Across different datasets and configurations, we observed consistent trends with minimal variation. Note that the standard deviations reported are between different runs with different target models, surrogate models, and data splits. Despite the existing randomness in query generations (model extraction) and randomized samples used for calibrating $\tau$ (membership inference), inner variability of results when the same models (e.g., $M$, $S_{\text{100K}}$) are attacked or evaluated with the same splits of the data (e.g., $D_A$, $D_N$) is far less than the reported deviations.   

All experiments were conducted on an Apple M2 Pro Mac Mini (A2816) with a 10-core CPU, 16-core GPU, 16-core Neural Engine, 16 GB of unified memory, and 1 TB of disk space. All machine learning models were trained using the CPU. The overall compute cost was modest, with each experiment completing in under a few hours. However, it is important to note that increasing the query budgets or the internal parameters of the membership inference attack (e.g., the number of adversarial samples generated) may lead to higher runtime and memory consumption. The total compute usage for the project, including preliminary and discarded trials, remained within the limits of a single-CPU setup and is reproducible on commodity hardware.

\section{Image Experiments Protocol and Training}
\label{app:image_protocol}

We conduct all image experiments on CIFAR-10~\cite{cifar}, a benchmark dataset consisting of 60{,}000 color images of size $32\times32$ evenly distributed across 10 classes. CIFAR-10 is officially partitioned into 50{,}000 training samples and 10{,}000 test samples. Following the SeqMIA protocol, we repartition CIFAR-10 into three disjoint subsets of 20{,}000 samples each for target training/testing, surrogate training/testing, and distillation. The target split is divided into 10{,}000 training samples and 10{,}000 test samples. The surrogate split is divided analogously into 10{,}000 training samples and 10{,}000 test samples. The final 20{,}000-sample split is reserved as a distillation dataset and is not used to train either the target or surrogate model directly. This split structure ensures that target and surrogate models are trained on disjoint data while remaining in-distribution (ID) with respect to CIFAR-10, and it provides a separate pool for distillation, enabling SeqMIA's sequential metric construction.

For ResNet-50, we train with SGD (learning rate $0.1$, momentum $0.9$, weight decay $5\times 10^{-4}$) for 100 epochs. We enable data augmentation to improve generalization. The learning rate is scheduled using MultiStepLR with milestones at epochs 60 and 80 and multiplicative decay factor $\gamma=0.1$. For VGG-16, we again train with SGD (learning rate $0.1$, momentum $0.9$, weight decay $5\times 10^{-4}$) for 100 epochs with augmentation enabled. We use StepLR with step size 5 and $\gamma=0.9$, so the learning rate is multiplied by $0.9$ every 5 epochs. These training recipes are applied consistently to both the target and the ID surrogate model.

\section{Auxiliary Seed Sensitivity}
\label{app:seed}

\begin{figure}[t]
\centering

\begin{subfigure}{0.48\columnwidth}
\centering
\includegraphics[width=\linewidth]{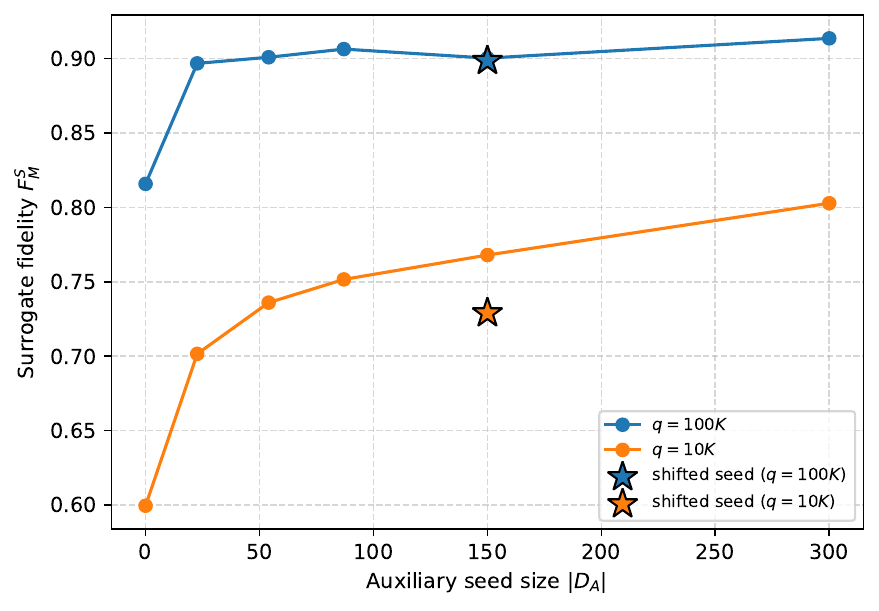}
\caption{Surrogate fidelity $F_M^S$.}
\label{fig:da_fidelity}
\end{subfigure}
\hfill
\begin{subfigure}{0.48\columnwidth}
\centering
\includegraphics[width=\linewidth]{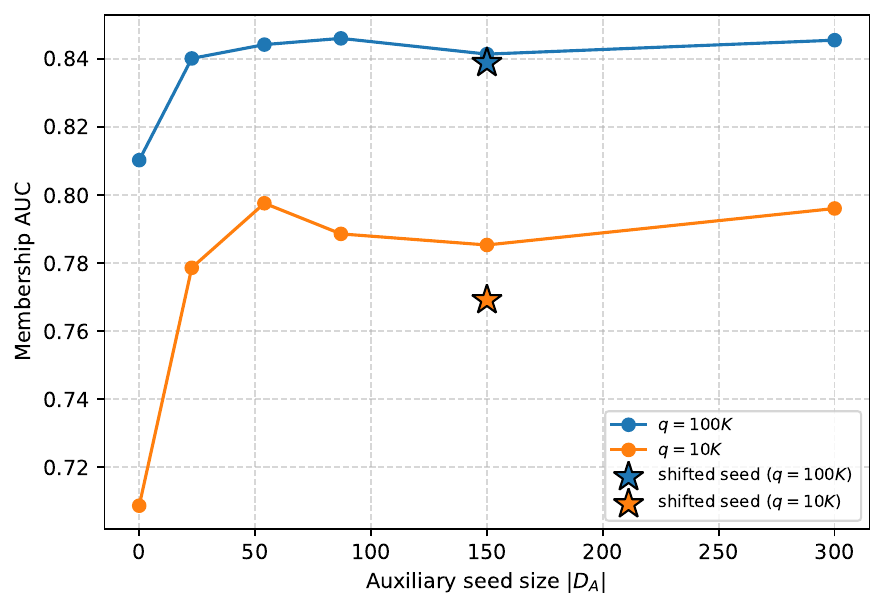}
\caption{Membership AUC.}
\label{fig:da_auc}
\end{subfigure}

\caption{Effect of the auxiliary seed size $D_A$ on surrogate fidelity and membership inference performance.}
\label{fig:da_sensitivity}
\end{figure}

We vary the size of seed set $D_A$ on the Location dataset to characterize the role from a near data-free setting ($|D_A| = 0$, where extraction is driven entirely by synthetic queries) up to $|D_A| = 300$. We additionally evaluate a distribution-shifted seed of the baseline size drawn from a class-imbalanced subpopulation. For each configuration we report surrogate fidelity $F_M^S$ and membership AUC at two budgets, $q = 10\text{K}$ and $q = 100\text{K}$, averaged over three runs with independently trained targets. Figures~\ref{fig:da_fidelity} and~\ref{fig:da_auc} summarize the results.

The attack is essentially insensitive to the seed at the larger seed. From the smallest real seed to $|D_A| = 300$, fidelity stays near $0.90$ and AUC near $0.84$, varying by at most about two points across the entire range. Even the data-free seed remains close to this plateau, reaching $F_M^S = 0.82$ and AUC $=0.81$. Hence, once the query budget is sufficient, the surrogate is determined far more by querying than by the seed it was bootstrapped from. At the smaller budget the seed matters, but as a convergence accelerant rather than a ceiling. With $q = 10\text{K}$, fidelity rises monotonically with seed size, from $0.60$ in the data-free case to $0.80$ at $|D_A| = 300$, and AUC climbs steeply from $0.71$ to roughly $0.79$ once even a small real seed is available. The budget gap narrows accordingly: the AUC difference between $q = 10\text{K}$ and $q = 100\text{K}$ shrinks from about ten points in the data-free case to about five points at $|D_A| = 300$. A larger seed therefore substitutes for queries when the budget is tight, and its advantage disappears once the budget is ample.

The distribution-shifted seed follows the same logic. At $q = 10\text{K}$ it underperforms the matched seed of equal size (AUC $0.77$ versus $0.79$, fidelity $0.73$ versus $0.77$), but at $q = 100\text{K}$ the two are indistinguishable ($\text{AUC} \approx 0.84$ and $F_M^S \approx 0.90$ for both). Accordingly, both a smaller seed and a mismatched seed slow convergence rather than cap performance, and additional querying absorbs either handicap.

These results support our characterization of $D_A$ as a bootstrap rather than a standing requirement (Section~\ref{sec:systhreatmodel}). Its size and provenance govern how quickly the surrogate converges, not the leakage the attack ultimately achieves, so the dependence on auxiliary data is amortized away by the same extraction that amortizes the per-sample inference cost.

\section{Linear-Axis ROC Curves}
\label{app:roc_linear}

\begin{figure*}[t]
    \centering
    \begin{subfigure}[t]{0.32\textwidth}
        \centering
        \includegraphics[width=\textwidth]{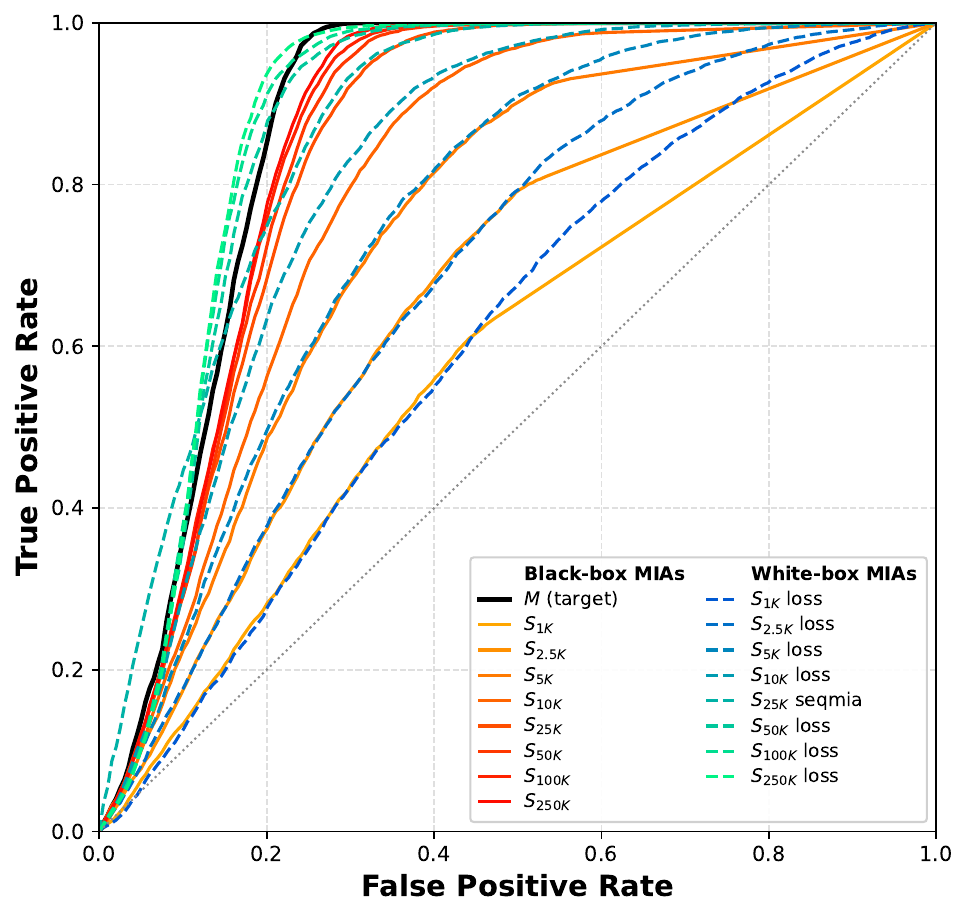}
        \caption{Location}
        \label{fig:roc_location_lin}
    \end{subfigure}
    \hfill
    \begin{subfigure}[t]{0.32\textwidth}
        \centering
        \includegraphics[width=\textwidth]{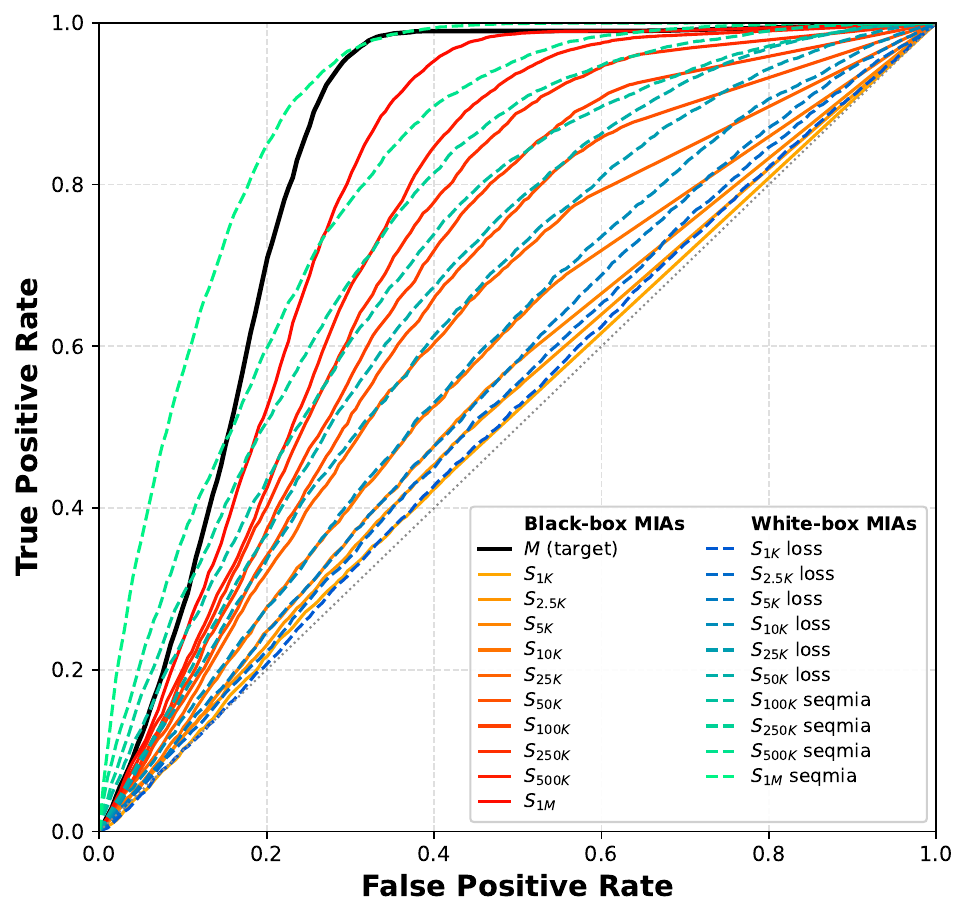}
        \caption{Purchase}
        \label{fig:roc_purchase_lin}
    \end{subfigure}
    \hfill
    \begin{subfigure}[t]{0.32\textwidth}
        \centering
        \includegraphics[width=\textwidth]{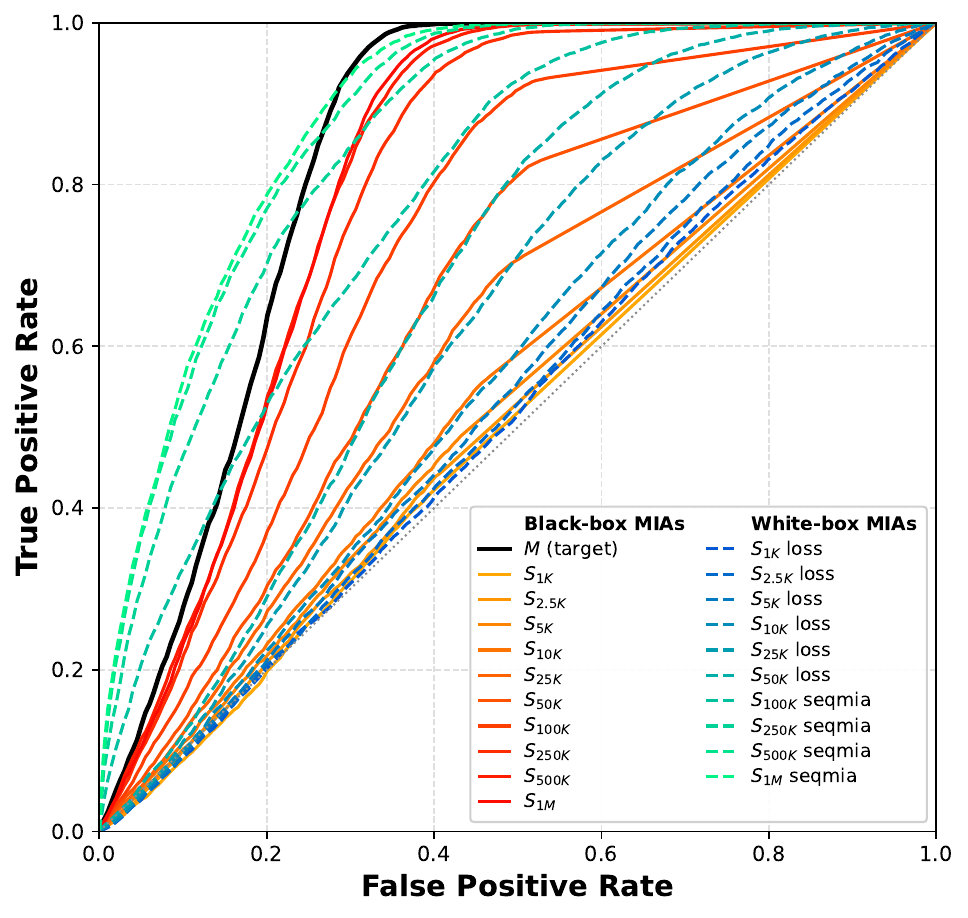}
        \caption{Texas}
        \label{fig:roc_texas_lin}
    \end{subfigure}
    \caption{Linear-axis ROC curves for membership inference on surrogate and target models across datasets.}
    \label{fig:roc_linear}
\end{figure*}

Figure~\ref{fig:roc_combined} presents the ROC curves on logarithmic axes, the view that resolves the low false-positive region where our white-box attacks separate from the target. Figure~\ref{fig:roc_linear} reproduces the same curves on linear axes.

\end{document}